\documentclass[letterpaper]{article} 
\usepackage{aaai25}  
\usepackage{times}  
\usepackage{helvet}  
\usepackage{courier}  
\usepackage[hyphens]{url}  
\usepackage{graphicx} 
\usepackage{natbib}  
\usepackage{caption} 
\usepackage{algorithm}
\usepackage{algorithmic}

\usepackage{comment}
\usepackage[utf8]{inputenc} 
\usepackage[T1]{fontenc}    
\usepackage{url}            
\usepackage{booktabs}       
\usepackage{amsfonts}       
\usepackage{nicefrac}       
\usepackage{microtype}      
\usepackage[table,xcdraw]{xcolor}         
\usepackage{graphicx}
\usepackage{amsmath}
\usepackage{dsfont}
\usepackage{multirow}
\usepackage{subcaption}
\usepackage{amssymb}

\usepackage{newfloat}
\usepackage{listings}
\DeclareCaptionStyle{ruled}{labelfont=normalfont,labelsep=colon,strut=off}
\floatstyle{ruled}
\newfloat{listing}{tb}{lst}{}
\floatname{listing}{Listing}
\title{Hierarchical Classification Auxiliary Network for Time Series Forecasting}
\author{
    Yanru Sun\textsuperscript{\rm 1}, Zongxia Xie\textsuperscript{\rm 1}\thanks{Corresponding author.},
    Dongyue Chen\textsuperscript{\rm 1}, Emadeldeen Eldele\textsuperscript{\rm 2,}\textsuperscript{\rm 3}, Qinghua Hu\textsuperscript{\rm 1}
}
\affiliations{
    \textsuperscript{\rm 1} Tianjin Key Lab of Machine Learning, College of Intelligence and Computing, Tianjin University, China \\

    \textsuperscript{\rm 2} Centre for Frontier AI Research, Agency for Science, Technology and Research, Singapore \\

    \textsuperscript{\rm 3} Institute for InfoComm Research, Agency for Science, Technology and Research, Singapore \\
    
    yanrusun@tju.edu.cn, caddiexie@hotmail.com, dyuechen@tju.edu.cn, emad0002@ntu.edu.sg, huqinghua@tju.edu.cn \\
}

\begin{document}

\maketitle

\begin{abstract}
Deep learning has significantly advanced time series forecasting through its powerful capacity to capture sequence relationships.
However, training these models with the Mean Square Error (MSE) loss often results in over-smooth predictions, making it challenging to handle the complexity and learn high-entropy features from time series data with high variability and unpredictability.
In this work, we introduce a novel approach by tokenizing time series values to train forecasting models via cross-entropy loss, while considering the continuous nature of time series data.
Specifically, we propose a \textbf{H}ierarchical \textbf{C}lassification \textbf{A}uxiliary \textbf{N}etwork, \textbf{HCAN}, a general model-agnostic component that can be integrated with any forecasting model.
HCAN is based on a Hierarchy-Aware Attention module that integrates multi-granularity high-entropy features at different hierarchy levels. At each level, we assign a class label for timesteps to train an Uncertainty-Aware Classifier. This classifier mitigates the over-confidence in softmax loss via evidence theory. We also implement a Hierarchical Consistency Loss to maintain prediction consistency across hierarchy levels.
Extensive experiments integrating HCAN with state-of-the-art forecasting models demonstrate substantial improvements over baselines on several real-world datasets.
\end{abstract}

\begin{links}
\link{Code}{https://github.com/syrGitHub/HCAN}
\end{links}

\section{Introduction}
Time series forecasting has received significant attention due to its wide-ranging social impact. Among existing approaches for time series forecasting, deep learning methods have emerged as significant contributors to this field \cite{zhou2021informer,zhou2022fedformer,zeng2023transformers,ni2024basisformer}. These methods showed a powerful capacity to capture sequence continuity features \cite{DBLP:conf/ijcai/WenZZCMY023,wang2024card} and enhance forecasting performance in practical applications such as finance \cite{hou2022multi}, weather forecasting \cite{lam2022graphcast}, resource planning \cite{chen2021graph}, and other domains \cite{shao2024exploring,wu2024catch}. 

\begin{figure}[t!]
    \centering
    \includegraphics[width=0.45\textwidth]{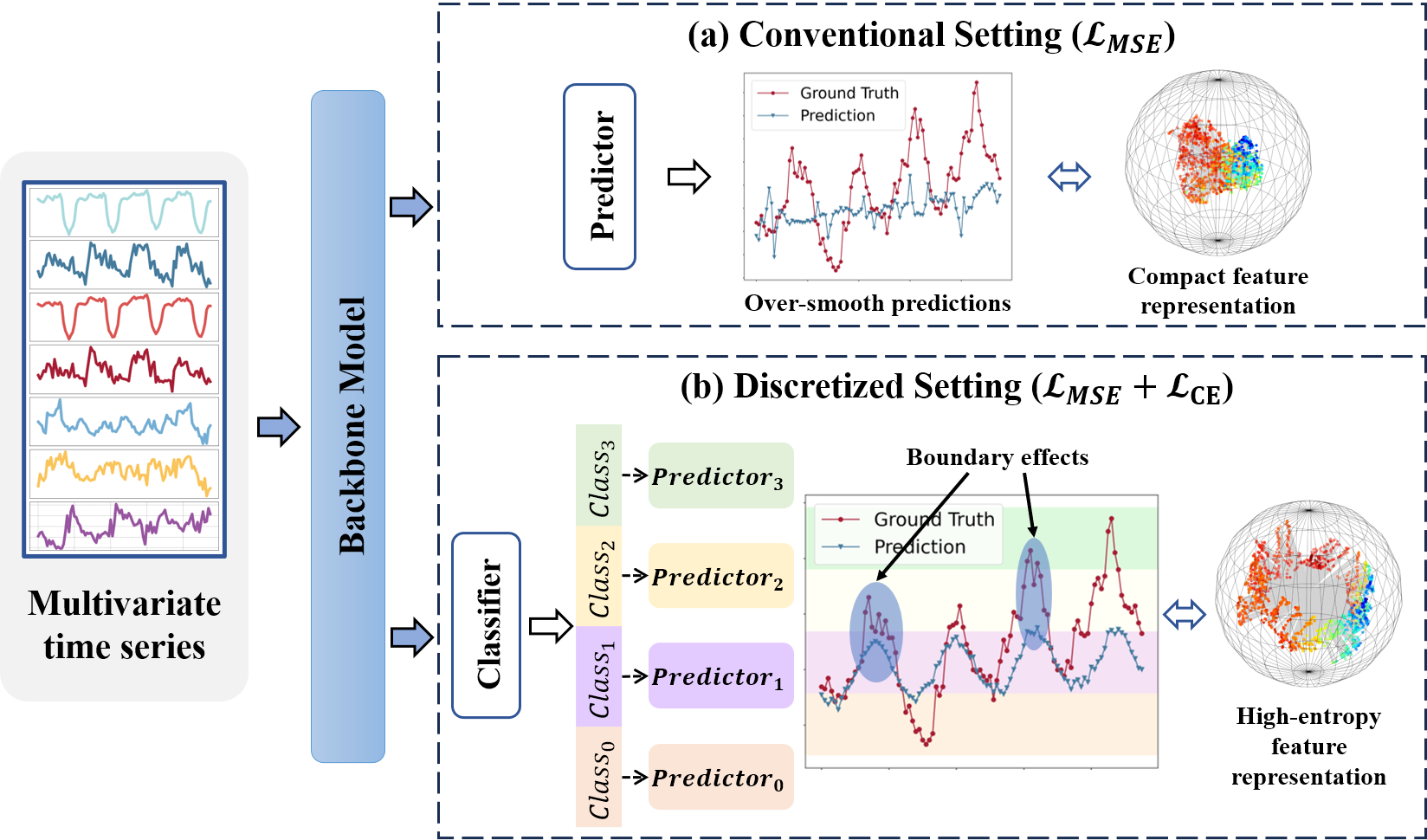}
    \caption{Comparison between Conventional and Discretized Settings for time series forecasting. (a) Conventional setting keeps features close together, producing over-smooth predictions; (b) Discretized setting spreads the features, resulting in a higher entropy feature space, but can misclassify inter-class boundary timesteps.}
    \label{fig:1}
\end{figure}

Nevertheless, current time series forecasting methods relying on the Mean Square Error (MSE) loss for feature extraction can suffer inaccurate predictions. The main downside of the MSE loss is compressing the feature representation into a narrow space, limiting its ability to learn complexity and high-entropy feature representations, especially for those features that exhibit significant variability and unpredictability \cite{zhang2023improving,pintea2023step}. 
Therefore, current methods often produce over-smooth predictions, leading to inaccuracies such as inflating wind speed estimates on sunny days when the actual wind speed is low, and underestimating wind speed on windy days when the actual wind speed is high.
This weakness diminishes the utility of forecasting results for downstream applications, as shown in Figure~\ref{fig:1}a.

Recently, several studies have demonstrated the superiority of cross-entropy loss in capturing high-entropy feature representation from a mutual information perspective \cite{pintea2023step,zhang2023improving}. Therefore, it has been successfully applied in various domains, such as depth estimation \cite{cao2017estimating,fu2018deep}, age estimation \cite{rothe2015dex,shah2024ordinal}, and crowd counting \cite{xiong2022discrete,guo2024consistency}.

In this work, we reformulate time series forecasting as a classification problem. Specifically, we tokenize time series values into different categories based on their magnitude and leverage the cross-entropy loss to train a classifier on these tokenized values. For example, in Figure~\ref{fig:1}b, we employ quantization to convert the real values into four discrete intervals, where each interval is considered a separate class. 
In this way, we can generate predictions within the corresponding interval based on the output of the classifier.

However, the continuous nature of time series data makes it challenging to classify values near the inter-class boundaries accurately. This difficulty may result in sub-optimal relative improvements, as illustrated by the blue circle in Figure~\ref{fig:1}b, a phenomenon commonly referred to as the \textit{boundary effects} \cite{liu2020facial}.

Therefore, we propose \textbf{H}ierarchical \textbf{C}lassification \textbf{A}uxiliary \textbf{N}etworks (\textbf{HCAN}), a novel model-agnostic component that can be integrated with any forecasting model. The architecture of HCAN is illustrated in Figure~\ref{fig:Model body}. In specific, we develop a Hierarchy-Aware Attention (HAA) module to incorporate multi-granular high-entropy features into the main features generated by the encoder network. For each hierarchy level, we propose an Uncertainty-Aware Classifier (UAC), combined with the evidence theory to mitigate the overconfident predictions and enhance the reliability of the features. Last, we propose a Hierarchical Consistency Loss (HCL) to ensure consistency of predicted values between hierarchies.
In summary, our contributions are as follows:

\begin{itemize}
    \item We reformulate forecasting as a hierarchical classification problem to introduce high-entropy feature representations, which helps to reduce over-smooth predictions.
    \item We propose HCAN, a hierarchy-aware attention module supported by uncertainty-aware classifiers and a consistency loss to alleviate issues caused by the boundary effects during the classification of timesteps.
    \item Extensive experiments conducted on real-world datasets show the effectiveness of integrating HCAN with various state-of-the-art methods.
\end{itemize}

\section{Related Work}
\subsection{Time Series Forecasting}
With the increased data availability and computational power, deep learning-based models have become an efficient solution to time series forecasting task \cite{qiu2024tfb}. In overall, based on the underlying network architecture, they can be categorized into models based on Recurrent neural networks (RNNs), Convolutional neural networks (CNNs), Transformer, and multi-layer perceptron (MLP). RNNs are traditionally utilized to capture temporal dependencies, yet they suffer from gradient vanishing and exploding problems.
In addition, besides the sequential data processing, RNNs have short-term memory and may not be efficient in learning long-term dependencies. 
To overcome the limitations of RNNs, Transformer-based models have excelled recently \cite{zhou2021informer,zhou2022fedformer,yu2023dsformer,liu2023itransformer}. Unlike RNNs, Transformers can process entire sequences simultaneously, benefiting from the parallel computations. In addition, Transformers handle long-range dependencies more effectively than RNNs \cite{nie2022time}.

On the other hand, recent studies have leveraged the robust abilities of CNNs to capture short-term patterns while attempting to enhance their capabilities for recognizing long-range dependencies \cite{liu2022scinet,tslanet}. Lastly, the recent development of MLP-based models has resulted in good performance with simple architectures \cite{zeng2023transformers,xu2024fits}.

Despite these advancements, these methods still struggle with capturing high-entropy feature representations due to their reliance on the MSE loss, which often leads to over-smooth predictions \cite{zhang2023improving}. Differently, our proposed work aims to overcome this limitation and construct a complex and high-entropy feature space, thereby enhancing feature diversity and improving prediction accuracy.

\begin{figure*}[t!]
    \centering
    \includegraphics[width=0.95\textwidth]{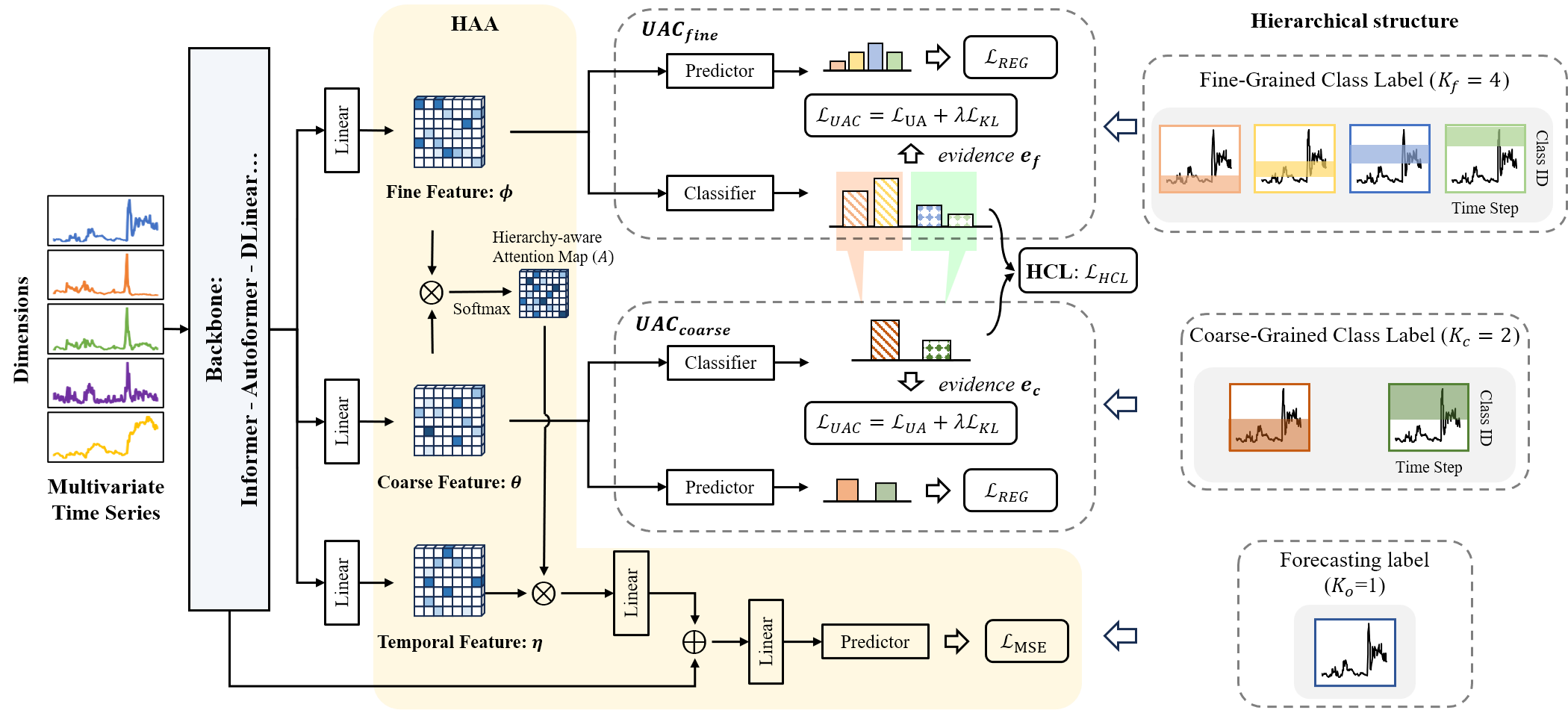}
    \caption{The structure of our proposed HCAN. From right to left, time series are first divided into fine-grained classes and coarse-grained classes  to form category labels for \textit{Hierarchical Classification}. According to these category labels, the \textit{Uncertainty-Aware Classifier} (UAC) at each level obtains reliable multi-granularity high-entropy features using evidence theory. The \textit{Hierarchical Consistency Loss} (HCL) ensures the consistency of values between hierarchies. Finally, the \textit{Hierarchy-Aware Attention} (HAA) module integrated the multi-granularity features into the forecasting features obtained by the backbones.}
    \label{fig:Model body}
\end{figure*}

\subsection{Classification for Continuous Targets}
Our approach draws inspiration from successful applications of classification in other domains, such as computer vision and pose estimation, where discretizing continuous targets has led to significant improvements \cite{rabanser2020effectiveness,gu2021dive}. For instance, in-depth estimation tasks, classifying depth ranges has proven more effective than precise value prediction \cite{cao2017estimating,fu2018deep}.

In the context of time series analysis, some recent works have explored limited categorization schemes. For example, DEMM \cite{wilson2022beyond} and DEMMA \cite{wang2023generalized} propose frameworks that segment time series into three broad categories. Similarly, NEC+ \cite{li2023extreme} employs binary classification to distinguish between extreme and normal events.

Our work significantly extends and refines these initial explorations by introducing a comprehensive, multi-level classification framework specifically designed for time series forecasting. This novel approach achieves a balance between the simplification benefits of discretization and the need for nuanced, continuous predictions. In addition, it addresses key limitations of previous methods, such as the loss of granularity in predictions and the occurrence of boundary effects near class thresholds.

\section{Methodology}
\subsection{Preliminaries}
\label{sec:preliminaries}

Given the historical time series data $X=\{x^i\}_{i=1}^N$ with $N$ samples, where $x^i\in \mathbb{R}^{L\times D}$, the goal of time series forecasting is to predict horizon series $Y=\{y^i\}_{i=1}^N$, where $y^i \in \mathbb{R}^{T\times D}$. Here, $L$ is the look-back window, $T$ is the number of future timesteps, and $D$ refers to the number of channels in the multivariate time series.

HCAN reformulates the forecasting task as a hierarchical classification task with 3 levels: the original series, coarse, and fine-grained. The number of categories at each level is $K_o=1$, $K_c=2$, $K_f=4$. 
At each level, a discretizing mapping function converts the continuous target $y^i$ into a categorical target $k^i$ based on which interval $\mathcal{I}_k = (\rho_k^{\text{left}}, \rho_k^{\text{right}})$ the value $y^i$ falls into. 
This interval $\mathcal{I}_k$ represents the range within which $y^i$ is categorized. 
The detailed mapping process can be found in the Appendix.
Subsequently, the relative forecasting target $\Delta y^i = y^i - \rho_k^{\text{left}}$ is computed as the offset of $y^i$ from the lower bound $\rho_k^{\text{left}}$ of the interval $\mathcal{I}_k$, where $\Delta y^i \in \mathbb{R}^{T\times D}$.
Therefore, the new structure of the dataset becomes $D = \{x^i, y^i, \Delta y^i_c, k^i_c, \Delta y^i_f, k^i_f\}_{i=1}^{N}$ with $N$ samples.

\subsection{Hierarchical Classification Auxiliary Network}
\label{sec:HCAN}
We propose a hierarchical structure that trains classifiers at the fine-grained and coarse-grained levels, each with a different number of classes, to obtain high-entropy features represented in multiple granularities. Specifically, the fine-grained feature is obtained from the hierarchy, which has a larger number of categories, providing the model with relatively precise quantification. Conversely, the coarse-grained feature, which corresponds to a hierarchy with fewer categories, aims to enhance classification accuracy, as shown in Figure~\ref{fig:Model body}. 

To illustrate the workflow of our HCAN, we begin by extracting features $F \in \mathbb{R}^{D \times T}$ from the backbone model.
Subsequently, we employ three distinct linear layers to generate three types of features: $\theta \in \mathbb{R} ^{D \times M}$, $\phi \in \mathbb{R} ^{D \times M}$, and $\eta \in \mathbb{R} ^{D \times M}$, representing fine-grained, coarse-grained, and the original temporal features, respectively.
Meanwhile, as depicted in the right-most part of Figure~\ref{fig:Model body}, we categorize the timesteps into fine-grained and coarse-grained classes based on their magnitude. Specifically, we define the boundary of each group by arranging the time series values in an ascending order and then dividing them based on the number of groups $K$ (see the Appendix). This categorization forms a hierarchical structure and establishes the category labels.

These hierarchical categories are used as labels to train the Uncertainty-Aware Classifiers (UAC) at the coarse-grained and fine-grained levels. Through backpropagation, the UAC refines the features $\theta$ and $\phi$, transforming them into high-entropy feature representations. The temporal feature $\eta$ is tailored to capture the temporal characteristics of time series forecasting.
Furthermore, we implement the Hierarchical Consistency Loss (HCL) to maintain consistency between the coarse-grained and fine-grained levels and to mitigate boundary effects.
Finally, we combine $\theta$, $\phi$, and $\eta$ with the initial forecasting features $F$ through the Hierarchy-Aware Attention (HAA) module. In the subsequent sections, we provide a detailed description of these components.

\subsubsection{Uncertainty-Aware Classification}
In our HCAN, we include a classifier at the coarse-grained and fine-grained levels to create the high entropy features. However, a key challenge is the high confidence often erroneously assigned to incorrect predictions by traditional softmax-based classifiers \cite{moon2020confidence,van2020uncertainty}. This issue becomes more obvious given our objective of classifying timesteps-level values into distinct classes. To address this issue and improve the robustness of classification across various hierarchical levels, we implement an evidence-based uncertainty estimation technique, which is meant to enhance the precision of uncertainty assessments. Moreover, we consider the case of challenging samples that are usually estimated with high uncertainty by the Evidential Deep Learning (EDL) methods \cite{han2022trusted}. To prioritize these samples, we propose a novel uncertainty-aware loss function. This loss increases the importance of these challenging samples in the learning process. Essentially, if the sample is hard to classify, it helps the model recognize its difficulty and pays more attention to it.

Our approach utilizes an evidence-based uncertainty estimation technique, leveraging the parameters of the Dirichlet distribution, which is the conjugate prior of the categorical distribution. This method allows us to compute belief masses ($b$) for different categories and the overall uncertainty mass ($u$), derived from the evidence ($e$) collected from the data.

For the $K$-class classification problems, the softmax layer of a conventional neural network classifier is replaced with an activation function layer (\textit{i.e.,} Softplus) to ensure non-negative outputs, which are then treated as evidence vectors $e\in \mathbb{R}_+^K$.
These vectors are obtained by the classifier network based on the fine-grained feature $\theta$ or coarse-grained feature $\phi$.
Next, we use these evidence vectors to construct the parameters of the Dirichlet distribution, $i.e.$, $\alpha = e + 1$, and calculate the belief mass $b_k$ and uncertainty $u$ as:
\begin{equation}
b_k=\frac{e_k}{S} = \frac{\alpha_k - 1}{S} \quad \text{and}\quad u = \frac{K}{S}, 
\label{eq:b_k,u}
\end{equation}
where $S=\sum_{i=1}^K(e_i+1) = \sum_{i=1}^K\alpha_i$ represents the Dirichlet strength. In addition, the sum of uncertainty mass $u$ and belief mass $b$ equals 1, $u + \sum_{k=1}^K b_k = 1$, where $u\geqslant 0$ and $b\geqslant 0$. Finally, the probability distribution $p$ is calculated as $p_k = \frac{\alpha_k}{S}$.

According to Eq.~\ref{eq:b_k,u}, the more evidence observed for the $k$-th class, the greater the probability allocated to the $k$-th class. Conversely, the less total evidence observed, the greater the overall uncertainty. Therefore, we use the belief mass to calculate the class uncertainty for each instance. Specifically, for the $i$-th sample, we use $(1-b^i)$ as class-level uncertainty, which is the uncertainty weight for categories during training. We define the uncertainty-aware (UA) coefficient as: $\omega^i = (1-b^i)\bigodot o^i$, where $\bigodot$ means the Hadamard product.

Finally, the UAC loss is defined as:
\begin{align}
\begin{split}
\mathcal{L}_{UAC} &= \lambda_{UA} \mathcal{L}^i_{UA} + \lambda_{KL} \mathcal{L}^i_{KL}\\
&= \lambda_{UA} \sum_{k=1}^K\omega_k^i(\psi (S^i) - \psi (\alpha_k^i))\\
&+\lambda_{KL} KL [Dir (p^i|\widetilde{\alpha}^i) || Dir (p^i|1)],
\end{split}
\end{align}
where $\psi(\cdot)$ is the digamma function, and $\lambda_{UA}, \lambda_{KL}$ are balance factors, and $\text{Dir} (p^i|1)$ approximates the uniform distribution. Notably, we make adjustments to the Dirichlet parameters $\alpha^i$ by  $\widetilde{\alpha}^i = o^i +(1-o^i)\bigodot \alpha^i$ to remove the non-misleading evidence.

By formalizing forecasting as a classification task, we introduce high entropy features into the forecasting feature space. At the same time, to encourage the continuity of the extracted features, we propose a relative prediction strategy, making predictions within each classification bin \cite{yu2021group}. 
We optimize using the MSE loss against the ground truth forecasting interval:
\begin{equation}
\mathcal{L}_{REG} = \sum_{k=1}^K \mathds{I}(c_k = 1)(\Delta y_k-\Delta \hat{y}_k)^2,
\end{equation}
where $c_k$ and $\Delta y_k$ denote the classification and relative prediction labels, respectively, and $\Delta \hat{y}_k$ is the relative prediction value obtained by the model.

The hierarchy loss is formulated across two layers with varying granularity as: 
\begin{equation}
\mathcal{L}_{HIER} = \mathcal{L}_{UAC}^f + \alpha \mathcal{L}_{REG}^f + \mathcal{L}_{UAC}^c + \alpha \mathcal{L}_{REG}^c,
\end{equation}
where $\alpha$ is the balance factor.

\subsubsection{Hierarchical Consistency Loss}  
\begin{figure}[t]
    \centering
    \includegraphics[width=0.45\textwidth]{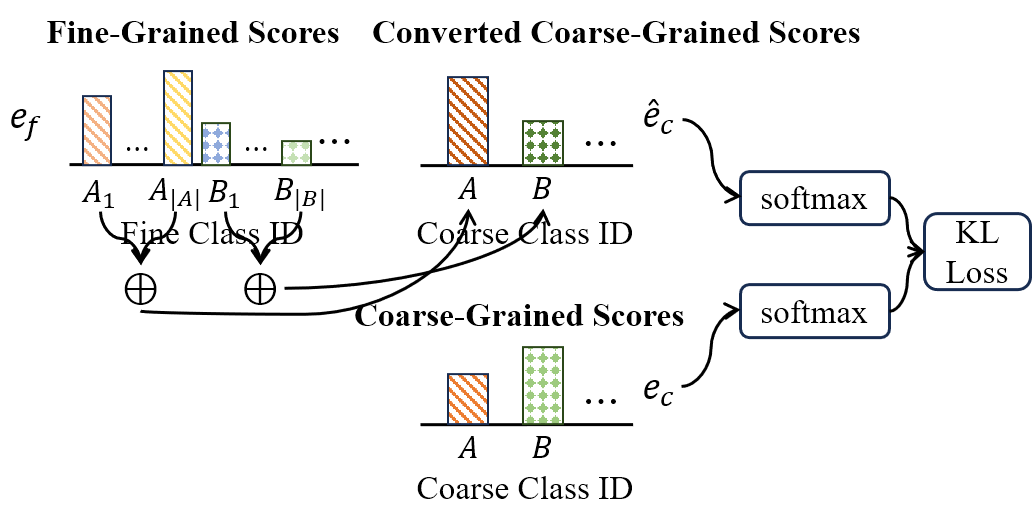}
    \caption{The hierarchical consistency loss between fine-grained and coarse-grained hierarchies encourages consistent predictions among them, alleviating the boundary effects. The $e_f$ from the fine-grained classifier is converted to $\hat{e}_c$, which aligns with the coarse-grained classifier $e_c$. We minimize the KL divergence loss between their softmax outputs.}
    \label{fig:HCL}
\end{figure}

Due to the continuous nature of time series data, directly classifying timestep values may result in misclassified values near the inter-class boundaries, known as \textit{boundary effects}. Therefore, we propose the Hierarchical Consistency Loss (HCL), which aims to keep the values near the boundary of a fine-grained class within the correct coarse-grained category.

To reinforce this alignment between the hierarchical classifiers, we propose an HCL to penalize discrepancies between them. As illustrated in Figure~\ref{fig:HCL}, we minimize a symmetric version of the Kullback-Leibler (KL) divergence between the class distributions of the fine-grained and coarse-grained classifiers.

For each fine-grained category, represented by evidence $e_f = [e_f^{A_1}, ..., e_f^{A_{|A|}}, e_f^{B_1}, ..., e_f^{B_{|B|}}, ...]$, we first convert it to a coarse-grained category evidence $e_c = [e_c^A, e_c^B,...]$. To align $e_f$ and $e_c$, we average the $e_f$ values that belong to the same coarse-grained class to produce the converted coarse-grained evidence: 

\begin{align}
\begin{split}
\hat{e}_c &= [\hat{e}_c^A, \hat{e}_c^B, \dots] \\
&= [\frac{e_f^{A_1} + ... + e_f^{A_{|A|}}}{|A|}, \frac{e_f^{B_1} + ... + e_f^{B_{|B|}}}{|B|}, ...].
\end{split}
\end{align}

The consistency loss for each coarse-grained class is then defined as a symmetric version of the KL divergence (equivalent to the Jensen-Shannon divergence) between $e$ and $\hat{e}$:
\begin{equation}
\mathcal{L}_{HCL} = \frac{1}{2}D_{KL}(e_c||\hat{e}_c) + \frac{1}{2}D_{KL}(\hat{e}_c||e_c). 
\end{equation}
This approach ensures that our model's predictions remain consistent across different hierarchical levels, effectively alleviating boundary effects.

\begin{table*}[!t]
\centering
\setlength{\tabcolsep}{3pt} 
\renewcommand{\arraystretch}{1.1} 
\resizebox{1.0\textwidth}{!}{
\begin{tabular}{cl|cc|cc|cc|cc|cc|cc|cc}
\toprule
\multicolumn{2}{c|}{Model}       & Informer & +HCAN          & Autoformer & +HCAN          & PatchTST       & +HCAN          & SCINet & +HCAN          & Dlinear & +HCAN          & iTransformer  & +HCAN          & FITS           & +HCAN          \\
\multicolumn{2}{c|}{Metric}      & MSE      & MSE            & MSE        & MSE            & MSE            & MSE            & MSE    & MSE            & MSE     & MSE            & MSE            & MSE            & MSE            & MSE            \\ \midrule
\multicolumn{2}{c|}{ETTh1}       & 1.077    & \textbf{0.897} & 0.530      & \textbf{0.462} & 0.421          & \textbf{0.396} & 0.591  & \textbf{0.536} & 0.453   & \textbf{0.428} & 0.457          & \textbf{0.451} & 0.439          & \textbf{0.436} \\
\multicolumn{2}{c|}{ETTh2}       & 4.779    & \textbf{2.359} & 0.483      & \textbf{0.406} & \textbf{0.342} & 0.343          & 1.041  & \textbf{0.820} & 0.473   & \textbf{0.411} & 0.384          & \textbf{0.375} & 0.375          & \textbf{0.368} \\
\multicolumn{2}{c|}{ETTm1}       & 0.951    & \textbf{0.717} & 0.606      & \textbf{0.540} & 0.353          & \textbf{0.350} & 0.417  & \textbf{0.390} & 0.359   & \textbf{0.344} & 0.408          & \textbf{0.403} & 0.414          & \textbf{0.405} \\
\multicolumn{2}{c|}{ETTm2}       & 1.729    & \textbf{0.981} & 0.359      & \textbf{0.303} & 0.258          & \textbf{0.250} & 0.753  & \textbf{0.685} & \textbf{0.287}   & 0.296 & 0.292          & \textbf{0.285} & 0.286          & \textbf{0.280} \\
\multicolumn{2}{c|}{Weather}     & 0.733    & \textbf{0.370} & 0.351      & \textbf{0.303} & 0.268          & \textbf{0.254} & 0.242  & \textbf{0.225} & 0.247   & \textbf{0.237} & 0.260          & \textbf{0.250} & 0.249          & \textbf{0.248} \\
\multicolumn{2}{c|}{Exchange}    & 1.726    & \textbf{0.845} & 0.525      & \textbf{0.410} & 0.516          & \textbf{0.344} & 0.844  & \textbf{0.549} & 0.369   & \textbf{0.338} & \textbf{0.364} & 0.395          & \textbf{0.360} & 0.426          \\
\multicolumn{2}{c|}{ILI}         & 2.889    & \textbf{2.738} & 5.012      & \textbf{4.166} & 1.516          & \textbf{1.428} & 3.277  & \textbf{3.265} & 2.347   & \textbf{2.276} & 2.767          & \textbf{2.741} & 3.680          & \textbf{2.095} \\
\multicolumn{2}{c|}{Electricity} & 0.352    & \textbf{0.337} & 0.250      & \textbf{0.236} & 0.259          & \textbf{0.233} & 0.213  & \textbf{0.209} & 0.210   & \textbf{0.208} & 0.176          & \textbf{0.167} & 0.217          & \textbf{0.216} \\
\multicolumn{2}{c|}{Traffic}     & 0.853    & \textbf{0.818} & 0.651      & \textbf{0.552} & 0.490          & \textbf{0.460} & 0.612  & \textbf{0.527} & 0.625   & \textbf{0.597} & 0.422          & \textbf{0.416} & 0.642          & \textbf{0.624} \\
\multicolumn{2}{c|}{Solar Wind}  & 1.953    & \textbf{1.025} & 1.362      & \textbf{1.057} & 1.109          & \textbf{0.948} & 1.174  & \textbf{1.091} & 1.071   & \textbf{1.019} & 1.360          & \textbf{1.028} & 1.349          & \textbf{1.239} \\ \bottomrule
\end{tabular}}
\caption{Multivariate long sequence time-series forecasting results. We report the MSE of different prediction lengths. The look-up window is set to $L = 336$ for PatchTST, DLinear, and SCINet, and $L = 96$ for other models. The \textbf{best results} are highlighted in \textbf{bold}. Detailed results of all prediction lengths for MSE/MAE are provided in the Appendix.}
\label{tab:SOTA Multivariate Quantitative Results}
\end{table*}

\subsubsection{Hierarchy-Aware Attention}
To introduce the high-entropy feature into the forecasting features to alleviate the over-smooth predictions, and optimize the trade-off between forecasting features and high-entropy features at different granularities, we have developed the Hierarchy-Aware Attention (HAA) module.

Building on the feature architecture of Hierarchical Classification Auxiliary Network, we reshape $\phi \in \mathbb{R}^{H \times D}$ projections, allowing their dot-products to interact and generate the HAA map $A$ of size $\mathbb{R}^{D\times D}$. 
This is combined with $F$ through a residual connection to introduce high-entropy feature representations. The overall HAA process is defined as follows:
\begin{equation}
\begin{split}
&\hat{Y} = W_f(W \cdot \text{Attention} (\theta, \phi, \eta) + F) + b, \\
&\text{Attention} (\theta, \phi, \eta) = \eta \cdot \text{Softmax} (\theta \cdot \phi),
\end{split}
\end{equation}
where $W$ and $W_f$ are linear layers, $F$ is the backbone feature map, and $\hat{Y}$ is the prediction output. 
The MSE loss is optimized according to $\hat{Y}$ and the ground truth labels $Y$ as:
\begin{equation}
\mathcal{L}_{MSE} = \frac{1}{N} \sum_{i=1}^N (Y^i - \hat{Y}^i)^2,
\end{equation}
where $N$ represents the number of samples.

To sum up, the overall training loss is defined as:
\begin{equation}
\begin{split}
\mathcal{L} = \mathcal{L}_{HIER} + \beta \mathcal{L}_{HCL} + \gamma \mathcal{L}_{MSE},
\end{split}
\end{equation}
where $\beta$ and $\gamma$ are hyper-parameter loss weights chosen through grid search.

\section{Experiments}
In this section, we conduct extensive experiments to evaluate the performance of HCAN and further perform ablation studies to justify how each component contributes to the results. Further details about the experimental setup can be found in the Appendix.

\subsection{Experimental Settings}

\paragraph{Datasets.}
We ran our experiments on ten publicly available real-world multivariate time series datasets, namely: \emph{ETT}, \emph{Exchange-Rate}, \emph{Weather}, \emph{ILI}, \emph{Electricity}, \emph{Traffic}, and \emph{Solar Wind}. We followed the standard protocol in the data preprocessing, where we split all datasets into training, validation, and testing in chronological order by a ratio of 6:2:2 for the ETT dataset and 7:1:2 for the other datasets \cite{zeng2023transformers}. See the Appendix for more details.

\paragraph{Backbone models.} 
We experimented our HCAN on top of several state-of-the-art deep learning-based forecasting models.
We selected these models with different architectures, where Informer \cite{zhou2021informer}, Autoformer \cite{wu2021autoformer}, PatchTST \cite{nie2022time}, and iTransformer \cite{liu2023itransformer} are Transformer-based models, SCINet \cite{liu2022scinet} is a CNN-based model, while DLinear \cite{zeng2023transformers} and FITS \cite{xu2024fits} are MLP-based models.
We evaluate their performance before and after including our HCAN in the multivariate and univariate settings. For the baselines, we re-run their codes in the same settings to ensure fairness and consistency.

\paragraph{Experiments details.}
Following previous works \cite{nie2022time, zeng2023transformers}, we used ADAM \cite{kingma2014adam} as the default optimizer across all the experiments and reported the MSE and mean absolute error (MAE) as the evaluation metrics. A lower MSE/MAE value indicates a better performance. Detailed results for MSE/MAE are provided in the Appendix.
We conducted the experiment for the same number of epochs as the baseline and the initial learning rate is chosen from \{5e-3, 1e-3, 5e-4, 1e-4, 5e-5, 1e-5\} through a grid search for different datasets. 
$\beta$ was chosen from \{1, 0.1, 0.01\} and $\gamma$ was chosen from \{1, 0.1, 0.01\} via grid search to obtain the best results.
For HCAN parameters, we set $K_c=2$ and $K_f=4$.
All the experiments were repeated five times with fixed random seeds, and we reported the average performance. HCAN was implemented by PyTorch \cite{paszke2019pytorch} and trained on a single NVIDIA RTX 3090 24GB GPU. 

\subsection{Main Results}

\begin{table*}[!t]
\centering\setlength{\tabcolsep}{3pt} 
\renewcommand{\arraystretch}{1.0} 
\resizebox{1.0\textwidth}{!}{
\begin{tabular}{cc|cc|cc|cc|cc|cc|cc|cc}
\toprule
\multicolumn{2}{c|}{Model}        & Informer & +HCAN          & Autoformer & +HCAN          & PatchTST       & +HCAN          & SCINet         & +HCAN          & Dlinear        & +HCAN          & iTransformer  & +HCAN          & FITS  & +HCAN          \\
\multicolumn{2}{c|}{Metric}       & MSE      & MSE            & MSE        & MSE            & MSE            & MSE            & MSE            & MSE            & MSE            & MSE            & MSE            & MSE            & MSE   & MSE            \\ \midrule
\multirow{4}{*}{\rotatebox{90}{ETTh1}}      & 96  & 0.255    & \textbf{0.121} & 0.088      & \textbf{0.082} & 0.055          & \textbf{0.055} & 0.088          & \textbf{0.068} & 0.057          & \textbf{0.053} & 0.061          & \textbf{0.060} & 0.056 & \textbf{0.054} \\
                            & 192 & 0.283    & \textbf{0.092} & 0.108      & \textbf{0.086} & \textbf{0.071} & 0.072          & 0.105          & \textbf{0.084} & 0.077          & \textbf{0.075} & 0.073          & \textbf{0.072} & 0.075 & \textbf{0.072} \\
                            & 336 & 0.291    & \textbf{0.088} & 0.118      & \textbf{0.091} & 0.082          & \textbf{0.078} & 0.130          & \textbf{0.094} & 0.097          & \textbf{0.088} & 0.089          & \textbf{0.087} & 0.091 & \textbf{0.089} \\
                            & 720 & 0.256    & \textbf{0.106} & 0.138      & \textbf{0.121} & 0.086          & \textbf{0.081} & 0.214          & \textbf{0.134} & 0.168          & \textbf{0.164} & \textbf{0.083} & 0.105          & 0.104 & \textbf{0.096} \\ \midrule
\multirow{4}{*}{\rotatebox{90}{ETTh2}}      & 96  & 0.302    & \textbf{0.182} & 0.169      & \textbf{0.140} & 0.129          & \textbf{0.127} & 0.130          & \textbf{0.129} & 0.133          & \textbf{0.128} & 0.135          & \textbf{0.133} & 0.125 & \textbf{0.123} \\
                            & 192 & 0.264    & \textbf{0.206} & 0.211      & \textbf{0.179} & 0.169          & \textbf{0.162} & 0.327          & \textbf{0.169} & 0.177          & \textbf{0.174} & 0.182          & \textbf{0.178} & 0.177 & \textbf{0.174} \\
                            & 336 & 0.324    & \textbf{0.223} & 0.255      & \textbf{0.226} & 0.187          & \textbf{0.187} & \textbf{0.198} & 0.220          & \textbf{0.212} & 0.225          & 0.218          & \textbf{0.215} & 0.222 & \textbf{0.221} \\
                            & 720 & 0.302    & \textbf{0.249} & 0.334      & \textbf{0.292} & 0.224          & \textbf{0.201} & 0.486          & \textbf{0.221} & 0.298          & \textbf{0.259} & 0.240          & \textbf{0.238} & 0.258 & \textbf{0.255} \\ \midrule
\multirow{4}{*}{\rotatebox{90}{ETTm1}}      & 96  & 0.093    & \textbf{0.046} & 0.059      & \textbf{0.047} & 0.026          & \textbf{0.024} & 0.049          & \textbf{0.029} & 0.030          & \textbf{0.026} & 0.029          & \textbf{0.028} & 0.029 & \textbf{0.027} \\
                            & 192 & 0.232    & \textbf{0.059} & 0.081      & \textbf{0.057} & 0.039          & \textbf{0.037} & 0.077          & \textbf{0.049} & 0.044          & \textbf{0.043} & 0.049          & \textbf{0.045} & 0.043 & \textbf{0.042} \\
                            & 336 & 0.271    & \textbf{0.108} & 0.088      & \textbf{0.072} & 0.053          & \textbf{0.050} & 0.109          & \textbf{0.089} & 0.064          & \textbf{0.059} & 0.061          & \textbf{0.060} & 0.057 & \textbf{0.056} \\
                            & 720 & 0.464    & \textbf{0.118} & 0.122      & \textbf{0.079} & 0.074          & \textbf{0.070} & 0.139          & \textbf{0.117} & \textbf{0.081} & 0.082          & 0.083          & \textbf{0.082} & 0.079 & \textbf{0.075} \\ \midrule
\multirow{4}{*}{\rotatebox{90}{ETTm2}}      & 96  & 0.092    & \textbf{0.065} & 0.127      & \textbf{0.095} & 0.065          & \textbf{0.065} & 0.079          & \textbf{0.069} & 0.064          & \textbf{0.061} & 0.069          & \textbf{0.069} & 0.070 & \textbf{0.069} \\
                            & 192 & 0.134    & \textbf{0.107} & 0.146      & \textbf{0.123} & 0.094          & \textbf{0.091} & 0.105          & \textbf{0.094} & 0.092          & \textbf{0.087} & 0.107          & \textbf{0.106} & 0.100 & \textbf{0.098} \\
                            & 336 & 0.178    & \textbf{0.141} & 0.217      & \textbf{0.126} & 0.120          & \textbf{0.117} & 0.130          & \textbf{0.128} & 0.129          & \textbf{0.120} & 0.144          & \textbf{0.143} & 0.128 & \textbf{0.126} \\
                            & 720 & 0.221    & \textbf{0.156} & 0.198      & \textbf{0.184} & 0.172          & \textbf{0.169} & 0.175          & \textbf{0.155} & \textbf{0.176} & 0.181          & \textbf{0.185} & 0.187          & 0.178 & \textbf{0.176} \\ \midrule
\multirow{4}{*}{\rotatebox{90}{Solar Wind}} & 96  & 1.443    & \textbf{1.268} & 2.316      & \textbf{1.289} & 1.021          & \textbf{0.851} & 1.518          & \textbf{1.366} & 1.316          & \textbf{1.223} & 1.727          & \textbf{1.266} & 1.669 & \textbf{1.658} \\
                            & 192 & 1.765    & \textbf{1.581} & 2.765      & \textbf{1.590} & 1.130          & \textbf{1.030} & 1.836          & \textbf{1.723} & 1.568          & \textbf{1.549} & 2.273          & \textbf{1.568} & 2.308 & \textbf{2.280} \\
                            & 336 & 1.849    & \textbf{1.740} & 2.783      & \textbf{1.715} & 1.137          & \textbf{1.098} & 1.853          & \textbf{1.746} & 1.686          & \textbf{1.671} & 2.370          & \textbf{1.714} & 2.355 & \textbf{2.327} \\
                            & 720 & 1.826    & \textbf{1.694} & 2.606      & \textbf{1.701} & 1.125          & \textbf{1.041} & 1.672          & \textbf{1.547} & 1.660          & \textbf{1.654} & 2.228          & \textbf{1.679} & 2.220 & \textbf{2.189} \\ \bottomrule
\end{tabular}}
\caption{Univariate long sequence time-series forecasting results on ETT full benchmark and Solar Wind dataset. We report the MSE of different prediction lengths $T \in \{96, 192, 336, 720\}$ for comparison. The look-up window is set to $L = 336$ for PatchTST, DLinear, and SCINet, and $L = 96$ for other models. The \textbf{best results} are highlighted in \textbf{bold}. Detailed results of all prediction lengths for MSE/MAE are provided in the Appendix.}
\label{tab:SOTA Univariate Quantitative Results}
\end{table*}

\begin{table*}[!t]
\centering
\centering\setlength{\tabcolsep}{3pt} 
\renewcommand{\arraystretch}{1.0} 
\resizebox{0.7\textwidth}{!}{
\begin{tabular}{ccccc|cccc|cccc}
\toprule
\multicolumn{5}{c|}{Component}                                                                        & \multicolumn{4}{c|}{Weather}                                                             & \multicolumn{4}{c}{Solar Wind}                                                           \\ \midrule
\multicolumn{2}{c}{$UAC_{fine}$} & \multirow{2}{*}{Hierarchy} & \multirow{2}{*}{$\mathcal{L}_{HCL}$} & \multirow{2}{*}{HAA} & \multirow{2}{*}{96} & \multirow{2}{*}{192} & \multirow{2}{*}{336} & \multirow{2}{*}{720} & \multirow{2}{*}{96} & \multirow{2}{*}{192} & \multirow{2}{*}{336} & \multirow{2}{*}{720} \\ \cline{1-2}
$\mathcal{L}_{UAC}$     & $\mathcal{L}_{REG}$     &                            &                         &                      &                     &                      &                      &                      &                     &                      &                      &                      \\ \midrule
-          & -          & -                          & -                       & -                    & 0.352               & 0.636                & 0.680                & 1.265                & 1.710               & 1.991                & 1.958                & 2.154                \\
\checkmark          & -          & -                          & -                       & -                    & 0.349               & 0.509                & 0.613                & 0.993                & 0.991               & 1.077                & 1.127                & 1.149                \\
\checkmark          & \checkmark          & -                          & -                       & -                    & 0.300               & 0.515                & 0.579                & 0.999                & 0.964               & 1.060                & 1.129                & 1.125                \\
\checkmark          & \checkmark          & \checkmark                          & -                       & -                    & 0.322               & 0.406                & 0.580                & 0.961                & 0.948               & 1.048                & 1.099                & 1.109                \\
\checkmark          & \checkmark          & \checkmark                          & \checkmark                      & -                    & 0.295               & 0.345                & 0.395                & 0.614                & 0.935               & 1.038                & 1.097                & 1.083                \\
\checkmark          & \checkmark          & \checkmark                          & \checkmark                       & \checkmark                    & \textbf{0.291}      & \textbf{0.306}       & \textbf{0.369}       & \textbf{0.513}       & \textbf{0.920}      & \textbf{1.027}       & \textbf{1.087}       & \textbf{1.065}       \\ \bottomrule
\end{tabular}}
\caption{Ablation study of the components of HCAN on the Weather and Solar Wind datasets using Informer as a backbone: Uncertainty-Aware Classification (UAC), Hierarchical Structure (Hierarchy), Hierarchical Consistency Loss ($\mathcal{L}_{HCL}$), and Hierarchy-Aware Attention (HAA). The results are in terms of MSE for different prediction lengths. The \textbf{best results} are highlighted in \textbf{bold.}}
\label{tab:Ablation Study}
\end{table*}
\textbf{Multivariate Forecasting Results.}
We present the multivariate forecasting results in Table~\ref{tab:SOTA Multivariate Quantitative Results}. Notably, our proposed HCAN demonstrates a substantial impact on the performance of the baselines, as it boosts their forecasting results by a noticeable margin. This is evident in 66 out of 70 cases. For instance, HCAN achieves average performance gains of 9.1\%, 35.5\%, 10.2\%, and 22.3\% on the ETT dataset series. Similar improvements also observed on other datasets.

We attribute these performance enhancements to two primary aspects. First, HCAN incorporates a reliable hierarchical classification structure that captures high-entropy features, effectively alleviating the over-smooth predictions and reducing the boundary discontinuity typically associated with classification tasks.
Second, the HAA mechanism enhances prediction accuracy by fusing features at different granular levels, thereby providing more reliable information for prediction. This attribute proves particularly advantageous in long-term forecasting scenarios, which inherently pose greater challenges as the forecast horizon extends. For example, as shown in the Appendix, when forecasting a length of 720 timesteps, the integration of HCAN with Autoformer leads to a significant reduction of 31.9\% in MSE on the ETTh2 dataset and a reduction of 19.3\% on the Exchange dataset. These results underscore the capability of HCAN to deliver stable and reliable predictions even in long-term forecasting scenarios.

\textbf{Univariate Forecasting Results.}
We also report the univariate forecasting outcomes for the ETT and Solar Wind datasets in Table~\ref{tab:SOTA Univariate Quantitative Results}. Compared to the original performance of the baseline methods, incorporating our HCAN into these models yields an overall reduction of 23.0\%, 35.8\%, 7.5\%, 12.6\%, 2.5\%, 22.8\%, and 1.5\% in the MSE results. These results validate the effectiveness of our proposed hierarchical structure in enhancing forecasting precision.

\subsection{Ablation Study}
\label{sec:Ablation Study}
Table~\ref{tab:Ablation Study} presents an ablation study on the Weather and Solar Wind datasets to assess the effectiveness of each module in HCAN. Referring to Figure~\ref{fig:Model body}, we evaluate the following settings: \textbf{(1)} including the UAC with only the fine-grained classes ($\mathcal{L}_{UAC}$ alone) \textbf{(2)} with adding $\mathcal{L}_{REG}$ to the UAC module, i.e., $\mathcal{L}_{UAC}+\mathcal{L}_{REG}$ \textbf{(3)} with including the coarse-grained classes and directly concatenating the multi-level features (Hierarchy) \textbf{(4)} with using $\mathcal{L}_{HCL}$ to keep consistency among hierarchy levels \textbf{(5)} with using the attention module for feature fusion instead of direct concatenation (HAA).

\textbf{Impact of UAC.} Initially, applying the UAC on the fine-grained features alone with $\mathcal{L}_{UAC}$ significantly enhances performance by creating a high-entropy feature space that enriches forecasting representations. Adding $\mathcal{L}_{REG}$ further improves performance by imposing relative forecasting constraints, ensuring feature continuity and coherence.

\textbf{Impact of Hierarchy Structure.} Implementing a hierarchical structure with two layers of UAC layers (by including the coarse-grained features) demonstrates the value of incorporating multi-granularity features, as indicated by performance gains in the ablation study.

\textbf{Impact of HCL.} Performance is further enhanced by integrating $\mathcal{L}_{HCL}$, which imposes a consistency constraint between hierarchies and effectively addresses boundary effects.

\textbf{Impact of HAA.} The best performance is observed when replacing direct concatenation with the HAA mechanism. This change indicates that different features contribute variably to forecasting outcomes, and simple concatenation can lead to sub-optimal results.

\begin{figure}[!t]
	\centering
	\begin{subfigure}[t]{0.14\textwidth} 
		\includegraphics[width=\textwidth]{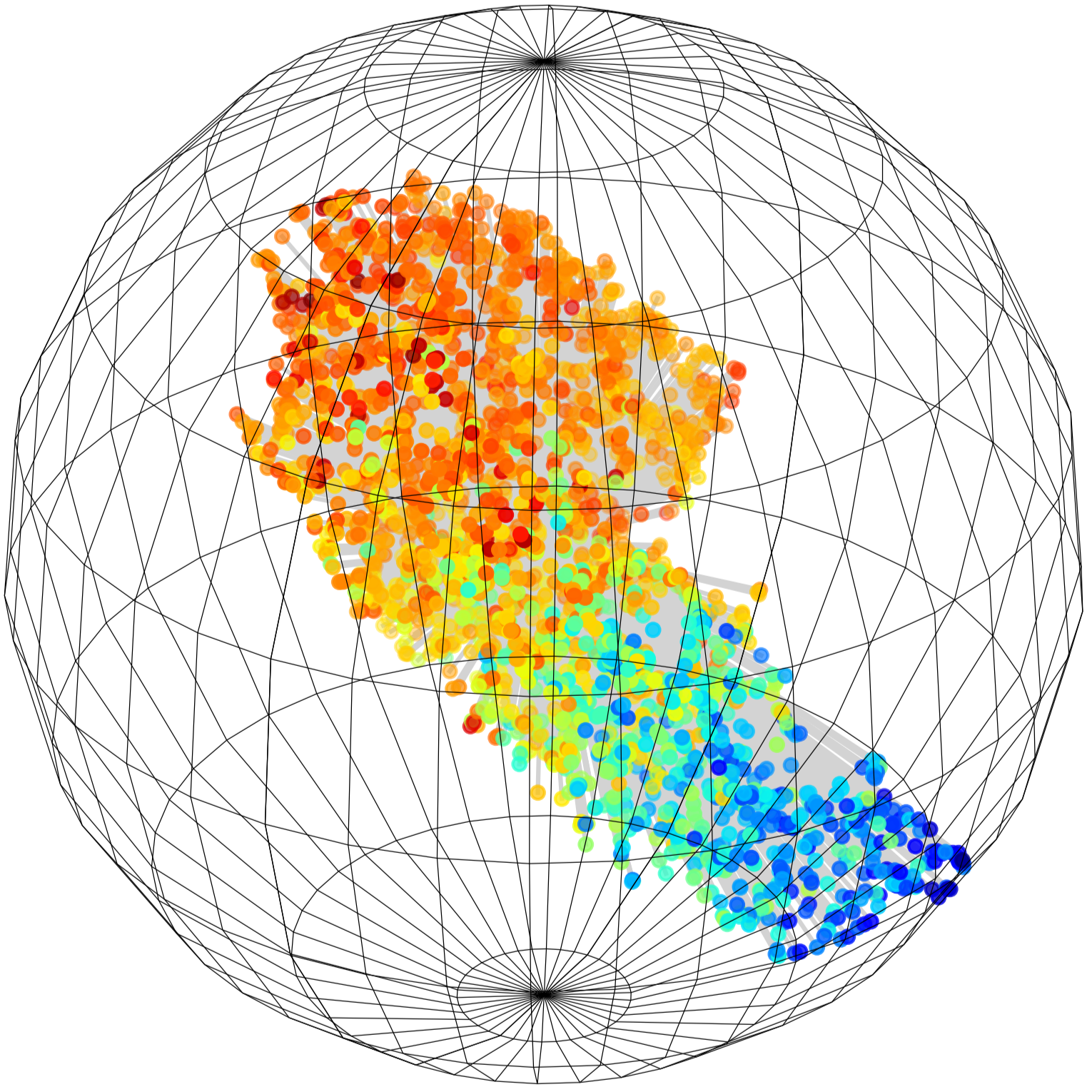} 
		\caption{SCINet}
		\label{fig:5-a}
	\end{subfigure}
	\hspace{1cm}
	\begin{subfigure}[t]{0.24\textwidth}
		\includegraphics[width=\textwidth]{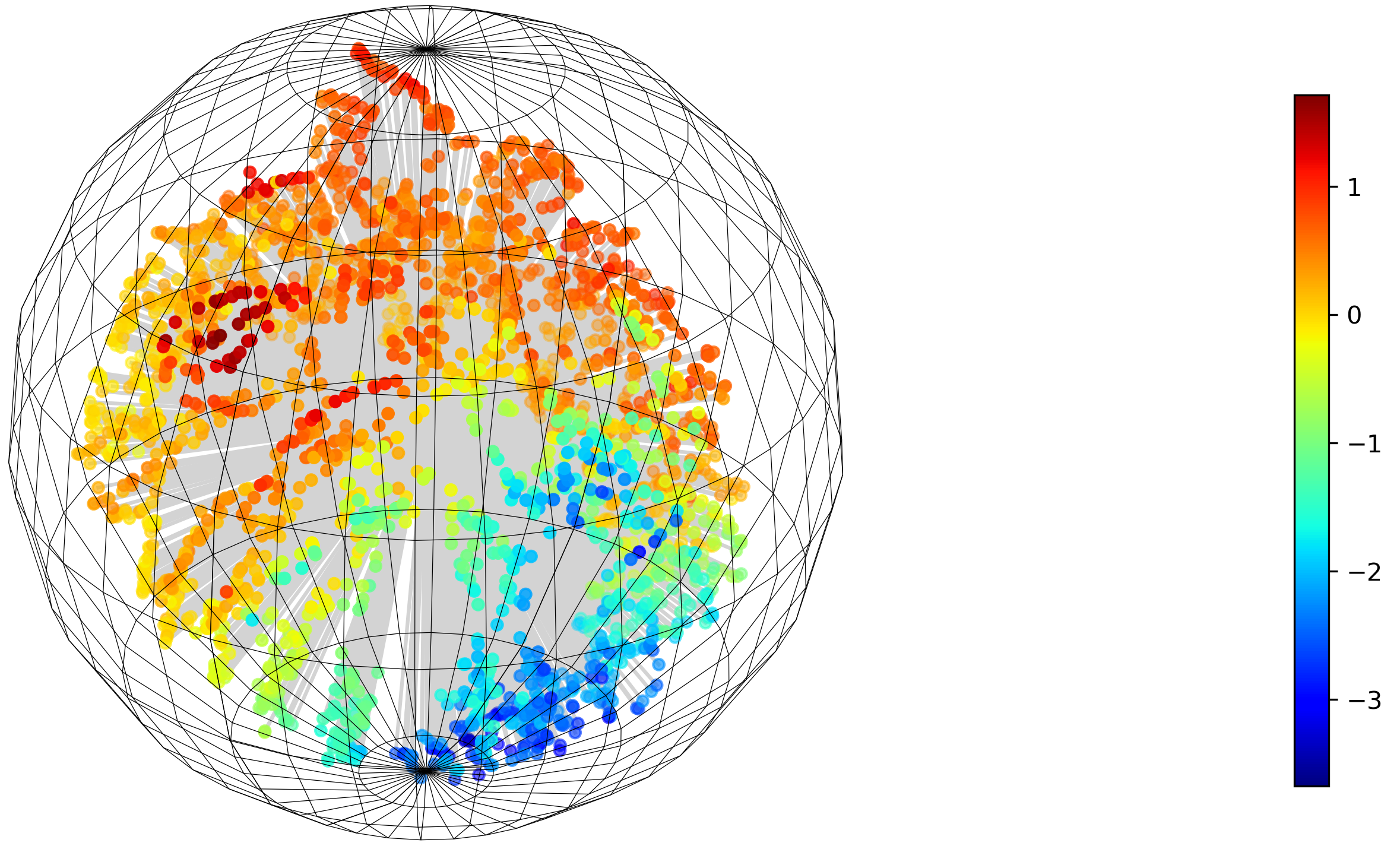} 
		\caption{Fine Feature}
		\label{fig:5-b}
	\end{subfigure}   
	\hspace{1cm} 
	\begin{subfigure}[t]{0.14\textwidth} 
		\includegraphics[width=\textwidth]{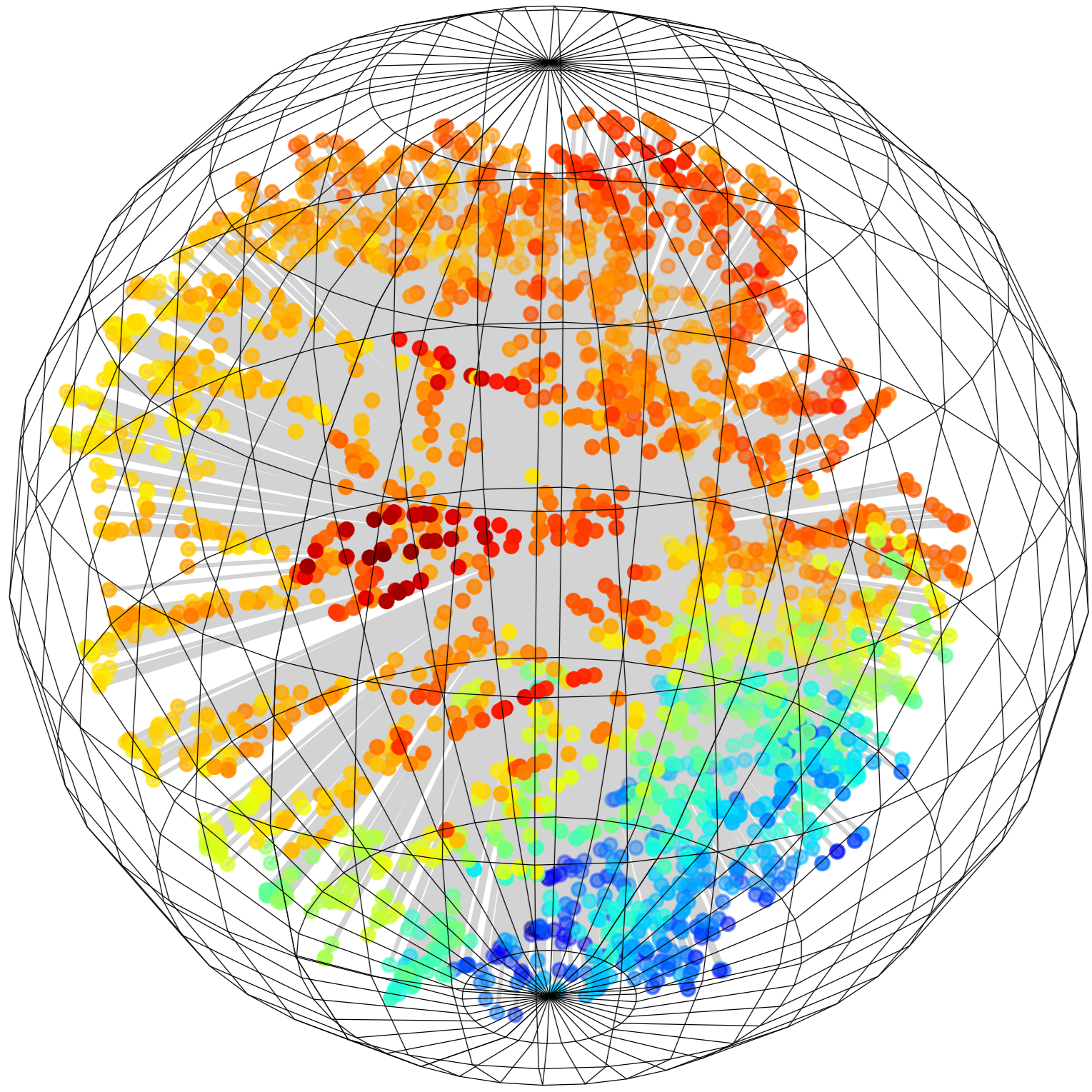} 
		\caption{Coarse Feature}
		\label{fig:5-c}
	\end{subfigure}
	\hspace{1cm}
	\begin{subfigure}[t]{0.24\textwidth}
		\includegraphics[width=\textwidth]{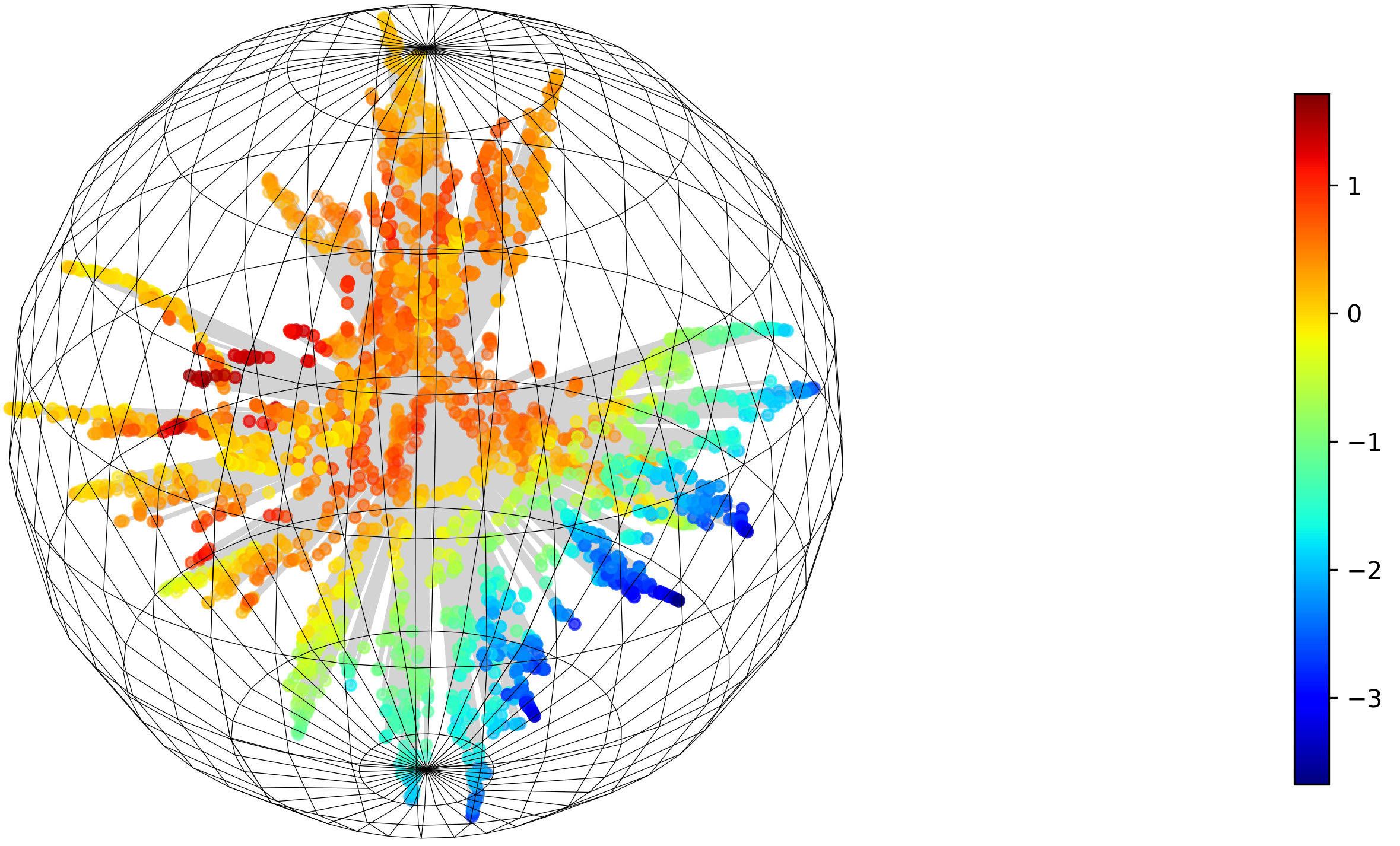} 
		\caption{Fusion Feature}
		\label{fig:5-d}
	\end{subfigure}
	\caption{t-SNE visualization of different features for SCINet on the ETTh1 dataset. (a) SCINet keeps features close together. (b)(c) Simply introducing classification spreads the features, obtaining a higher entropy feature space, while the ordinal relationship is lost. (d) By combining the classification features with the forecasting features, a high entropy and ordered feature representation is obtained. Features are coloured based on their predicted value.}
	\label{fig:t-SNE results}
\end{figure}

\begin{figure}[!t]
	\centering
    \begin{subfigure}[t]{0.20\textwidth} 
		\includegraphics[width=\textwidth]{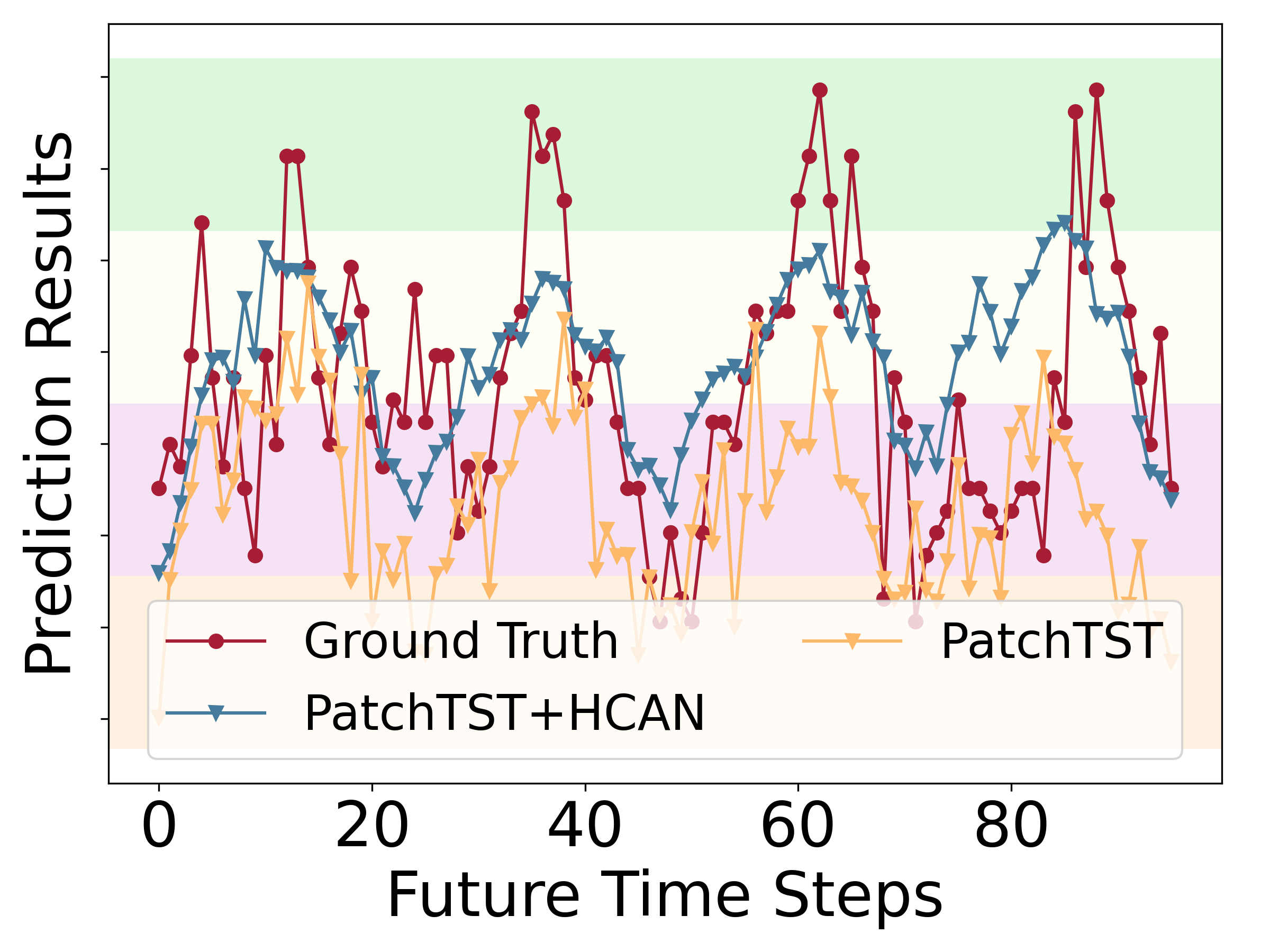} 
		\caption{PatchTST+HCAN}
		\label{fig:4-a}
    \end{subfigure}
	\hspace{1cm}
    \begin{subfigure}[t]{0.20\textwidth}
        \includegraphics[width=\textwidth]{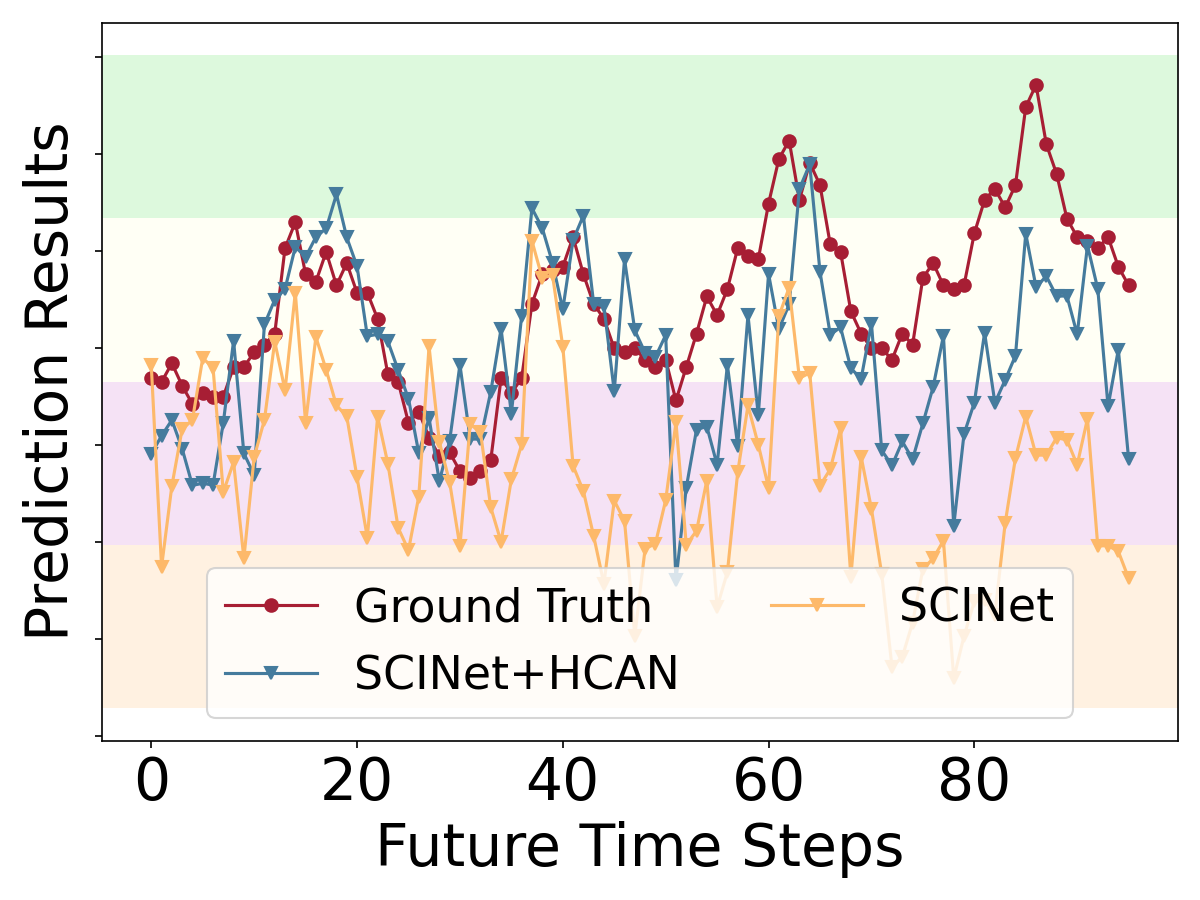} 
		\caption{SCINet+HCAN}
		\label{fig:4-b}
    \end{subfigure}
	\hspace{1cm}
    \begin{subfigure}[t]{0.20\textwidth}
        \includegraphics[width=\textwidth]{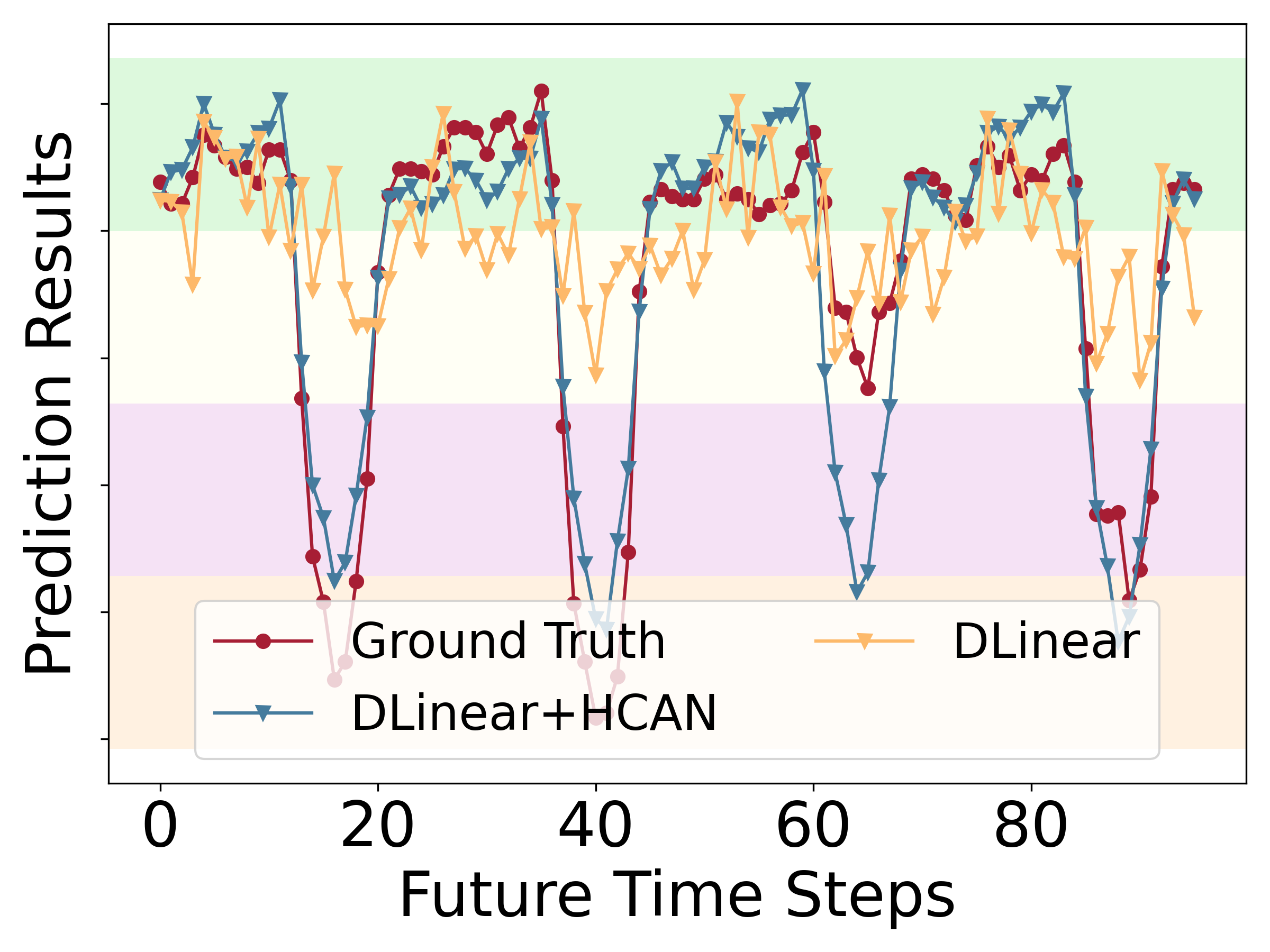} 
		\caption{DLinear+HCAN}
		\label{fig:4-c}
    \end{subfigure}
        \hspace{1cm}
    \begin{subfigure}[t]{0.20\textwidth}
        \includegraphics[width=\textwidth]{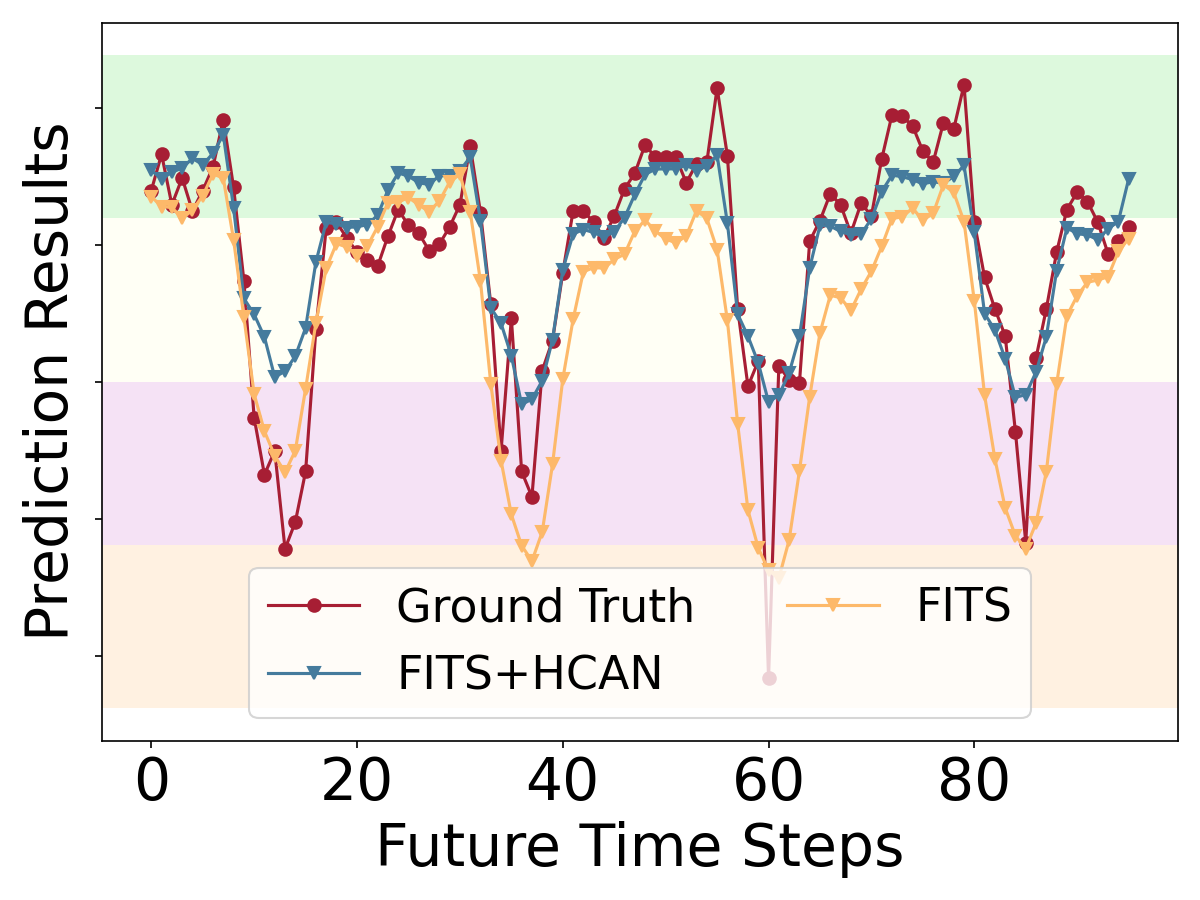} 
		\caption{FITS+HCAN}
		\label{fig:4-d}
    \end{subfigure}
	\caption{The prediction results (Horizon = 96) of (a) PatchTST vs. PatchTST+HCAN, (b) SCINet vs. SCINet+HCAN, (c) DLinear vs. DLinear+HCAN, (d) FITS vs. FITS+HCAN, on randomly-selected sequences from the ETTh1 dataset.}
	\label{fig:fitting results}
\end{figure}

\subsection{Qualitative Evaluation}
\label{sec:Qualitative Evaluation}
\textbf{High-entropy Feature Representation.} The t-SNE visualization of the features from SCINet on the ETTh1 dataset is displayed in Figure~\ref{fig:t-SNE results}. As depicted in Figure~\ref{fig:5-a}, representations learned from the MSE loss exhibit lower diversity. Figures~\ref{fig:5-b} and~\ref{fig:5-c} illustrate that integrating classification indeed spreads features more broadly, yet it disrupts ordinality in feature space. Figure~\ref{fig:5-d} shows how the HAA mechanism combines hierarchical features with the original features from the backbone model, effectively spreading the feature while maintaining ordinality. In conclusion, HCAN facilitates reliable high-entropy feature representations through hierarchical classification, significantly helping to alleviate over-smooth predictions. 

\textbf{Visualizations.} To examine the quality of prediction results with and without our HCAN, Figure~\ref{fig:fitting results} presents this comparison on PatchTST, SCINe, DLinear, and FITS backbones on the ETTh1 dataset.
Clearly, our HCAN yields more realistic predictions. This enhancement is largely regarded to the proposed hierarchical consistency loss (HCL), which notably improves performance at class boundaries. These results further validate the effectiveness of the high-entropy feature representations. Additionally, they demonstrate that HCL is effective in mitigating the boundary effects.

\section{Conclusion}
In this study, we addressed the issue of over-smooth predictions in time series forecasting by introducing a novel hierarchical classification from an entropy perspective.
We proposed HCAN, a model-agnostic component that enhances forecasting by tokenizing output and integrating muti-granularity high-entropy feature representations through a hierarchical-aware attention module. The HCL loss further aids in mitigating boundary effects, promoting overall accuracy.
Extensive experiments on benchmarking datasets demonstrate that HCAN substantially improves the performance of baseline forecasting models. Our results suggest that HCAN can serve as a foundation component in time series forecasting, providing deeper insights into the interplay between classification tasks and forecasting.

\section{Acknowledgments}
This work was supported in part by the National Natural Science Foundation of China under Grants 62376194, 61925602, U23B2049, 62406219, and 62436001 and in part by the China Postdoctoral Science Foundation - Tianjin Joint Support Program under Grant 2023T014TJ.

\bigskip
\bibliography{aaai25}

\clearpage
\appendix
\section{Datasets and Implementation Details}
\label{app:Dataset}
This subsection provides a summary of the datasets utilized in this paper: 
\begin{itemize}
    \item \textbf{ETT}~\footnote{\url{https://github.com/zhouhaoyi/ETDataset}} \cite{zhou2021informer} (Electricity Transformer Temperature) dataset contains two electric transformers, ETT1 and ETT2, collected from two separate counties. Each of them has two versions of sampling resolutions (15min \& 1h). Thus, there are four ETT datasets: \textbf{ETTm1}, \textbf{ETTm2}, \textbf{ETTh1}, and \textbf{ETTh2}. Oil temperature is the target series.
    \item \textbf{Weather}~\footnote{\url{https://www.bgc-jena.mpg.de/wetter/}} \cite{wu2021autoformer} dataset contains 21 meteorological indicators in Germany, such as humidity and air temperature. The $CO_2$ is chosen as the target series.
    \item \textbf{Exchange-Rate}~\footnote{\url{https://github.com/laiguokun/multivariate-time-series-data}}  \cite{lai2018modeling} the exchange-rate dataset contains the daily exchange rates of eight foreign countries including Australia, British, Canada, Switzerland, China, Japan, New Zealand, and Singapore ranging from 1990 to 2016. We consider the time series of 30 days as a sample for this task. The Singapore exchange is taken as the target series, and we aim to predict the exchange rate of Singapore each day of a month.
    \item \textbf{ILI}~\footnote{\url{https://gis.cdc.gov/grasp/fluview/fluportaldashboard.html}} \cite{wu2021autoformer} dataset collects the number of patients and influenza-like illness ratio in a weekly frequency. The "total patients" is chosen as the target series.
    \item \textbf{Electricity}~\footnote{\url{https://archive.ics.uci.edu/ml/datasets/ElectricityLoadDiagrams20112014}} \cite{wu2021autoformer} is a dataset that describes 321 customers’ hourly electricity consumption. The "320" is chosen as the target series.
    \item \textbf{Traffic}~\footnote{\url{http://pems.dot.ca.gov}} \cite{wu2021autoformer} is a dataset featuring hourly road occupancy rates from 862 sensors along the freeways in the San Francisco Bay area. The "861" is chosen as the target series.
    \item \textbf{Solar Wind}~\footnote{\url{https://github.com/syrGitHub/TDAM}} \cite{sun2021solar} dataset released by NASA is a collection of hourly solar wind properties from 2011 to 2017 collected by many spacecraft orbiting the L1 point between the Sun and Earth. The solar wind speed is the target series.
\end{itemize}

For data split, we follow \cite{zhou2021informer} and split data into train/validation/test set by the ratio 6:2:2 towards ETT datasets. We follow \cite{zeng2023transformers} to preprocess data and split data by the ratio of 7:1:2 in other datasets. Details are shown in Table~\ref{tab:Dataset}

\begin{table}[htbp]
	\centering
	\scalebox{0.7}{
		\begin{tabular}{c|cccc}
			\toprule
			Dataset         & Variates & Prediction Length     & Timesteps & Granularity \\ \midrule
			ETTh1 \& ETTh2  & 7        & \{96, 192, 336, 720\} & 17420     & 1 hour      \\
			ETTm1 \& ETTm2  & 7        & \{96, 192, 336, 720\} & 69680     & 5 min       \\
			Weather         & 21       & \{96, 192, 336, 720\} & 52696     & 10 min      \\
                Exchange-Rate   & 8        & \{96, 192, 336, 720\} & 7588      & 1 day    \\
			ILI             & 7        & \{24, 36, 48, 60\}    & 966       & 7 day      \\
			Electricity     & 321      & \{96, 192, 336, 720\} & 26304     & 1 hour      \\
                Traffic         & 862      & \{96, 192, 336, 720\} & 17544     & 1 hour   \\
			Solar-Wind      & 4        & \{96, 192, 336, 720\} & 61369     & 1 hour      \\ \bottomrule
	\end{tabular}}
	\caption{The statistics of the ten datasets.}
	\label{tab:Dataset}
\end{table}

\section{Group Mapping Strategy}
We describe our partition strategy to define the boundary of each group. First, for the list of time series $y = [y_1, ..., y_Q]$, where $Q$ is the length of the time series, we arrange them in ascending order to obtain $\hat{y} = [\hat{y}_1, ..., \hat{y}_Q]$. Given the number of groups $K$, the partitioning algorithm defines the boundary of each interval $\mathcal{I}_k = (\rho_k^{\text{left}}, \rho_k^{\text{right}})$ as follows:
\begin{align}
\begin{split}
&\rho_k^{\text{left}} = \hat{y}(\lfloor (Q-1)\times \frac{(k-1)}{K} \rfloor),\\
&\rho_k^{\text{right}} = \hat{y}(\lfloor (Q-1)\times \frac{k}{K} \rfloor), \quad \forall k=1,2,\dots, K,
\end{split}
\end{align}
where we use $\hat{y}(k)$ to represent the $k$-th element of y. It is worth noting that the group strategy is non-trivial. If we simply divide the entire range uniformly into multiple groups, the time series within some of these groups in the training set may be unbalanced.

\section{Related Work}
\subsection{Multi-scale Modeling for Time series}
Recently, multi-scale modeling has gained attention for its ability to capture temporal dependencies at different granularities, which is critical for time series forecasting. 
Pyraformer \cite{liu2022pyraformer} introduces a pyramid attention mechanism to extract features at various temporal resolutions, enabling models to capture patterns at different scales. 
Preformer \cite{du2023preformer} proposes multi-scale segment-wise correlations as an extension to the self-attention mechanism, enhancing the model's ability to understand complex temporal structures.
Scaleformer \cite{shabani2023scaleformer} proposes a multi-scale framework, and the need to allocate a predictive model at different temporal resolutions results in higher model complexity.
Similarly, TimesNet \cite{wu2023timesnet} ravels out the complex temporal variations into the multiple intraperiod- and intrerperiod-variations to adaptively discover the multi-periodicity within the data .
Pathformer \cite{chen2024pathformer} further advanced this concept by using a multi-scale Transformer with adaptive pathways to capture complex temporal relationships across scales. 
Timemixer \cite{wang2024timemixer} proposed a multi-scale mixing architecture, emphasizing that combining patterns from different scales improves forecasting accuracy.

Our work aligns with and builds upon these developments by integrating multi-scale modeling in a way that not only addresses the complexities of temporal dependencies but also enhances model flexibility across different scales.

\begin{table*}[!t]
\centering
\centering\setlength{\tabcolsep}{3pt} 
\renewcommand{\arraystretch}{1.5} 
\resizebox{1.0\textwidth}{!}{
\large
\begin{tabular}{cc|cccc|cccc|cccc|cccc|cccc|cccc|cccc}
\toprule
\multicolumn{2}{c|}{Model}                              & \multicolumn{2}{c}{Informer} & \multicolumn{2}{c|}{+HCAN}      & \multicolumn{2}{c}{Autoformer}  & \multicolumn{2}{c|}{+HCAN}      & \multicolumn{2}{c}{PatchTST}    & \multicolumn{2}{c|}{+HCAN}      & \multicolumn{2}{c}{SCINet}      & \multicolumn{2}{c|}{+HCAN}      & \multicolumn{2}{c}{Dlinear}     & \multicolumn{2}{c|}{+HCAN}      & \multicolumn{2}{c}{iTransforrmer} & \multicolumn{2}{c|}{+HCAN}      & \multicolumn{2}{c}{FITS}        & \multicolumn{2}{c}{+HCAN}       \\
\multicolumn{2}{c|}{Metric}                             & MSE           & MAE          & MSE            & MAE            & MSE            & MAE            & MSE            & MAE            & MSE            & MAE            & MSE            & MAE            & MSE            & MAE            & MSE            & MAE            & MSE            & MAE            & MSE            & MAE            & MSE             & MAE             & MSE            & MAE            & MSE            & MAE            & MSE            & MAE            \\ \midrule
\multicolumn{1}{c|}{\multirow{5}{*}{\rotatebox{90}{ETTh1}}}       & 96  & 0.950         & 0.773        & \textbf{0.703} & \textbf{0.617} & 0.465          & 0.459          & \textbf{0.412} & \textbf{0.405} & 0.382          & 0.405          & \textbf{0.358} & \textbf{0.398} & 0.445          & 0.460          & \textbf{0.415} & \textbf{0.422} & 0.384          & 0.405          & \textbf{0.371} & \textbf{0.382} & 0.387           & 0.405           & \textbf{0.379} & \textbf{0.402} & 0.385          & 0.393          & \textbf{0.377} & \textbf{0.388} \\
\multicolumn{1}{c|}{}                             & 192 & 1.011         & 0.787        & \textbf{0.848} & \textbf{0.704} & 0.484          & 0.471          & \textbf{0.441} & \textbf{0.444} & 0.414          & 0.421          & \textbf{0.382} & \textbf{0.412} & 0.457          & 0.459          & \textbf{0.421} & \textbf{0.420} & 0.443          & 0.450          & \textbf{0.420} & \textbf{0.392} & 0.441           & 0.436           & \textbf{0.432} & \textbf{0.427} & 0.435          & 0.422          & \textbf{0.429} & \textbf{0.417} \\
\multicolumn{1}{c|}{}                             & 336 & 1.141         & 0.844        & \textbf{0.979} & \textbf{0.767} & 0.517          & 0.495          & \textbf{0.490} & \textbf{0.459} & 0.437          & 0.443          & \textbf{0.421} & \textbf{0.432} & 0.573          & 0.541          & \textbf{0.556} & \textbf{0.532} & 0.451          & 0.451          & \textbf{0.439} & \textbf{0.414} & 0.491           & 0.462           & \textbf{0.489} & \textbf{0.454} & 0.475          & 0.443          & \textbf{0.481} & \textbf{0.436} \\
\multicolumn{1}{c|}{}                             & 720 & 1.207         & 0.864        & \textbf{1.058} & \textbf{0.800} & 0.655          & 0.573          & \textbf{0.505} & \textbf{0.504} & 0.450          & 0.466          & \textbf{0.425} & \textbf{0.451} & 0.888          & 0.705          & \textbf{0.754} & \textbf{0.674} & 0.535          & 0.536          & \textbf{0.482} & \textbf{0.491} & 0.509           & 0.494           & \textbf{0.504} & \textbf{0.474} & 0.462          & \textbf{0.460} & \textbf{0.456} & 0.475          \\ \cline{2-30} 
\rowcolor{pink!50}
\multicolumn{1}{c|}{\cellcolor{white}}                            & Avg & 1.077         & 0.817        & \textbf{0.897} & \textbf{0.722} & 0.530          & 0.500          & \textbf{0.462} & \textbf{0.453} & 0.421          & 0.434          & \textbf{0.396} & \textbf{0.423} & 0.591          & 0.541          & \textbf{0.536} & \textbf{0.512} & 0.453          & 0.460          & \textbf{0.428} & \textbf{0.420} & 0.457           & 0.449           & \textbf{0.451} & \textbf{0.439} & 0.439          & 0.430          & \textbf{0.436} & \textbf{0.429} \\ \midrule
\multicolumn{1}{c|}{\multirow{5}{*}{\rotatebox{90}{ETTh2}}}       & 96  & 2.896         & 1.342        & \textbf{1.563} & \textbf{1.011} & 0.383          & 0.412          & \textbf{0.341} & \textbf{0.364} & 0.275          & 0.337          & \textbf{0.265} & \textbf{0.334} & 0.746          & 0.637          & \textbf{0.648} & \textbf{0.600} & 0.300          & 0.363          & \textbf{0.287} & \textbf{0.358} & 0.301           & 0.350           & \textbf{0.282} & \textbf{0.343} & 0.290          & 0.339          & \textbf{0.284} & \textbf{0.322} \\
\multicolumn{1}{c|}{}                             & 192 & 6.580         & 2.117        & \textbf{2.757} & \textbf{1.463} & 0.463          & 0.463          & \textbf{0.408} & \textbf{0.404} & 0.339          & 0.379          & \textbf{0.323} & \textbf{0.363} & 0.860          & 0.689          & \textbf{0.716} & \textbf{0.642} & 0.394          & 0.426          & \textbf{0.359} & \textbf{0.400} & 0.380           & 0.399           & \textbf{0.373} & \textbf{0.381} & 0.377          & 0.391          & \textbf{0.372} & \textbf{0.382} \\
\multicolumn{1}{c|}{}                             & 336 & 5.608         & 1.994        & \textbf{2.734} & \textbf{1.435} & 0.473          & 0.474          & \textbf{0.460} & \textbf{0.468} & \textbf{0.331} & \textbf{0.380} & 0.368          & 0.401          & 1.000          & 0.744          & \textbf{0.764} & \textbf{0.688} & 0.465          & 0.471          & \textbf{0.439} & \textbf{0.444} & 0.424           & 0.432           & \textbf{0.420} & \textbf{0.426} & 0.416          & 0.425          & \textbf{0.408} & \textbf{0.419} \\
\multicolumn{1}{c|}{}                             & 720 & 4.034         & 1.673        & \textbf{2.384} & \textbf{1.332} & 0.614          & 0.527          & \textbf{0.418} & \textbf{0.450} & 0.421          & 0.494          & \textbf{0.416} & \textbf{0.440} & 1.557          & 0.954          & \textbf{1.153} & \textbf{0.863} & 0.733          & 0.606          & \textbf{0.557} & \textbf{0.534} & 0.430           & 0.447           & \textbf{0.423} & \textbf{0.435} & 0.418          & 0.437          & \textbf{0.409} & \textbf{0.421} \\ \cline{2-30} 
\rowcolor{pink!50}
\multicolumn{1}{c|}{\cellcolor{white}}                             & Avg & 4.779         & 1.782        & \textbf{2.359} & \textbf{1.310} & 0.483          & 0.469          & \textbf{0.406} & \textbf{0.422} & \textbf{0.342} & 0.397          & 0.343          & \textbf{0.384} & 1.041          & 0.756          & \textbf{0.820} & \textbf{0.698} & 0.473          & 0.467          & \textbf{0.411} & \textbf{0.434} & 0.384           & 0.407           & \textbf{0.375} & \textbf{0.396} & 0.375          & 0.398          & \textbf{0.368} & \textbf{0.386} \\ \midrule
\multicolumn{1}{c|}{\multirow{5}{*}{\rotatebox{90}{ETTm1}}}       & 96  & 0.670         & 0.595        & \textbf{0.592} & \textbf{0.544} & 0.534          & \textbf{0.490} & \textbf{0.489} & 0.491          & 0.289          & 0.343          & \textbf{0.281} & \textbf{0.329} & 0.394          & 0.414          & \textbf{0.345} & \textbf{0.390} & 0.301          & 0.345          & \textbf{0.284} & \textbf{0.321} & 0.342           & 0.377           & \textbf{0.339} & \textbf{0.360} & 0.354          & 0.375          & \textbf{0.336} & \textbf{0.370} \\
\multicolumn{1}{c|}{}                             & 192 & 0.855         & 0.702        & \textbf{0.620} & \textbf{0.573} & 0.595          & 0.511          & \textbf{0.515} & \textbf{0.497} & 0.336          & 0.371          & \textbf{0.318} & \textbf{0.342} & 0.385          & 0.422          & \textbf{0.362} & \textbf{0.404} & 0.336          & 0.366          & \textbf{0.328} & \textbf{0.349} & 0.383           & 0.396           & \textbf{0.379} & \textbf{0.388} & 0.392          & 0.393          & \textbf{0.383} & \textbf{0.395} \\
\multicolumn{1}{c|}{}                             & 336 & 1.149         & 0.827        & \textbf{0.721} & \textbf{0.621} & 0.683          & 0.552          & \textbf{0.563} & \textbf{0.536} & 0.367          & 0.392          & \textbf{0.349} & \textbf{0.351} & 0.408          & 0.430          & \textbf{0.402} & \textbf{0.422} & 0.372          & 0.389          & \textbf{0.351} & \textbf{0.372} & 0.418           & 0.418           & \textbf{0.414} & \textbf{0.403} & 0.425          & 0.415          & \textbf{0.408} & \textbf{0.410} \\
\multicolumn{1}{c|}{}                             & 720 & 1.129         & 0.786        & \textbf{0.935} & \textbf{0.716} & 0.614          & 0.527          & \textbf{0.590} & \textbf{0.471} & \textbf{0.419} & \textbf{0.425} & 0.452          & 0.432          & 0.479          & 0.471          & \textbf{0.451} & \textbf{0.458} & 0.427          & 0.423          & \textbf{0.413} & \textbf{0.421} & 0.487           & 0.457           & \textbf{0.482} & \textbf{0.440} & \textbf{0.486} & \textbf{0.449} & 0.492          & 0.454          \\ \cline{2-30} 
\rowcolor{pink!50}
\multicolumn{1}{c|}{\cellcolor{white}}                             & Avg & 0.951         & 0.728        & \textbf{0.717} & \textbf{0.613} & 0.606          & 0.520          & \textbf{0.540} & \textbf{0.499} & 0.353          & 0.382          & \textbf{0.350} & \textbf{0.364} & 0.417          & 0.434          & \textbf{0.390} & \textbf{0.418} & 0.359          & 0.381          & \textbf{0.344} & \textbf{0.366} & 0.408           & 0.412           & \textbf{0.403} & \textbf{0.398} & 0.414          & 0.408          & \textbf{0.405} & \textbf{0.407} \\ \midrule
\multicolumn{1}{c|}{\multirow{5}{*}{\rotatebox{90}{ETTm2}}}       & 96  & 0.447         & 0.523        & \textbf{0.419} & \textbf{0.505} & \textbf{0.243} & 0.324          & 0.248          & \textbf{0.291} & 0.164          & 0.254          & \textbf{0.161} & \textbf{0.242} & \textbf{0.208} & \textbf{0.304} & 0.225          & 0.325          & 0.172          & 0.267          & \textbf{0.169} & \textbf{0.264} & 0.186           & 0.272           & \textbf{0.183} & \textbf{0.264} & 0.183          & 0.266          & \textbf{0.181} & \textbf{0.260} \\
\multicolumn{1}{c|}{}                             & 192 & 0.814         & 0.706        & \textbf{0.592} & \textbf{0.621} & 0.284          & 0.341          & \textbf{0.253} & \textbf{0.327} & 0.224          & 0.294          & \textbf{0.218} & \textbf{0.293} & 0.351          & 0.410          & \textbf{0.307} & \textbf{0.387} & \textbf{0.237} & \textbf{0.314} & 0.279          & 0.395          & 0.254           & 0.314           & \textbf{0.242} & \textbf{0.312} & 0.247          & 0.305          & \textbf{0.231} & \textbf{0.290} \\
\multicolumn{1}{c|}{}                             & 336 & 1.426         & 0.916        & \textbf{0.927} & \textbf{0.735} & 0.366          & 0.390          & \textbf{0.295} & \textbf{0.353} & 0.278          & 0.330          & \textbf{0.276} & \textbf{0.330} & 0.608          & 0.548          & \textbf{0.601} & \textbf{0.511} & 0.307          & 0.358          & \textbf{0.306} & \textbf{0.350} & 0.316           & \textbf{0.351}  & \textbf{0.306} & 0.355          & 0.307          & 0.342          & \textbf{0.306} & \textbf{0.335} \\
\multicolumn{1}{c|}{}                             & 720 & 4.229         & 1.609        & \textbf{1.986} & \textbf{1.191} & 0.544          & 0.481          & \textbf{0.415} & \textbf{0.416} & 0.367          & 0.385          & \textbf{0.345} & \textbf{0.382} & 1.842          & \textbf{0.996} & \textbf{1.606} & 1.025          & 0.431          & 0.449          & \textbf{0.431} & \textbf{0.441} & 0.414           & 0.407           & \textbf{0.410} & \textbf{0.401} & 0.407          & 0.397          & \textbf{0.403} & \textbf{0.330} \\ \cline{2-30} 
\rowcolor{pink!50}
\multicolumn{1}{c|}{\cellcolor{white}}                             & Avg & 1.729         & 0.939        & \textbf{0.981} & \textbf{0.763} & 0.359          & 0.384          & \textbf{0.303} & \textbf{0.347} & 0.258          & 0.316          & \textbf{0.250} & \textbf{0.312} & 0.753          & 0.565          & \textbf{0.685} & \textbf{0.562} & \textbf{0.287} & \textbf{0.347} & 0.296          & 0.363          & 0.292           & 0.336           & \textbf{0.285} & \textbf{0.333} & 0.286          & 0.327          & \textbf{0.280} & \textbf{0.304} \\ \midrule
\multicolumn{1}{c|}{\multirow{5}{*}{\rotatebox{90}{Weather}}}     & 96  & 0.352         & 0.419        & \textbf{0.291} & \textbf{0.371} & 0.291          & 0.359          & \textbf{0.255} & \textbf{0.344} & 0.304          & 0.309          & \textbf{0.287} & \textbf{0.289} & 0.156          & 0.212          & \textbf{0.149} & \textbf{0.204} & 0.175          & 0.236          & \textbf{0.164} & \textbf{0.219} & 0.176           & 0.216           & \textbf{0.161} & \textbf{0.242} & 0.167          & 0.214          & \textbf{0.165} & \textbf{0.208} \\
\multicolumn{1}{c|}{}                             & 192 & 0.636         & 0.562        & \textbf{0.306} & \textbf{0.382} & 0.315          & 0.374          & \textbf{0.283} & \textbf{0.363} & 0.197          & 0.243          & \textbf{0.183} & \textbf{0.238} & 0.216          & 0.263          & \textbf{0.197} & \textbf{0.249} & 0.218          & 0.278          & \textbf{0.205} & \textbf{0.264} & 0.225           & 0.257           & \textbf{0.218} & \textbf{0.298} & 0.215          & 0.257          & \textbf{0.211} & \textbf{0.254} \\
\multicolumn{1}{c|}{}                             & 336 & 0.680         & 0.584        & \textbf{0.369} & \textbf{0.438} & 0.378          & 0.408          & \textbf{0.318} & \textbf{0.383} & 0.250          & 0.284          & \textbf{0.239} & \textbf{0.280} & 0.268          & 0.308          & \textbf{0.241} & \textbf{0.280} & 0.263          & 0.314          & \textbf{0.258} & \textbf{0.309} & 0.281           & \textbf{0.299}  & \textbf{0.276} & 0.347          & \textbf{0.267} & \textbf{0.293} & 0.270          & 0.295          \\
\multicolumn{1}{c|}{}                             & 720 & 1.265         & 0.815        & \textbf{0.513} & \textbf{0.545} & 0.423          & 0.431          & \textbf{0.356} & \textbf{0.396} & 0.320          & 0.334          & \textbf{0.305} & \textbf{0.320} & 0.329          & 0.351          & \textbf{0.312} & \textbf{0.344} & 0.332          & 0.374          & \textbf{0.319} & \textbf{0.357} & 0.358           & \textbf{0.350}  & \textbf{0.345} & 0.392          & 0.347          & 0.345          & \textbf{0.345} & \textbf{0.344} \\ \cline{2-30} 
\rowcolor{pink!50}
\multicolumn{1}{c|}{\cellcolor{white}}                             & Avg & 0.733         & 0.595        & \textbf{0.370} & \textbf{0.434} & 0.351          & 0.393          & \textbf{0.303} & \textbf{0.371} & 0.268          & 0.293          & \textbf{0.254} & \textbf{0.282} & 0.242          & 0.283          & \textbf{0.225} & \textbf{0.269} & 0.247          & 0.300          & \textbf{0.237} & \textbf{0.287} & 0.260           & \textbf{0.280}  & \textbf{0.250} & 0.320          & 0.249          & 0.277          & \textbf{0.248} & \textbf{0.275} \\ \midrule
\multicolumn{1}{c|}{\multirow{5}{*}{\rotatebox{90}{Exchange}}}    & 96  & 0.953         & 0.776        & \textbf{0.653} & \textbf{0.667} & 0.150          & \textbf{0.281} & \textbf{0.147} & 0.308          & 0.090          & 0.211          & \textbf{0.081} & \textbf{0.194} & 0.405          & 0.461          & \textbf{0.123} & \textbf{0.264} & 0.085          & 0.209          & \textbf{0.078} & \textbf{0.197} & 0.086           & 0.206           & \textbf{0.084} & \textbf{0.204} & 0.088          & 0.210          & \textbf{0.086} & \textbf{0.210} \\
\multicolumn{1}{c|}{}                             & 192 & 1.238         & 0.880        & \textbf{0.731} & \textbf{0.717} & 0.298          & 0.398          & \textbf{0.229} & \textbf{0.300} & 0.199          & 0.318          & \textbf{0.173} & \textbf{0.247} & 0.569          & 0.550          & \textbf{0.269} & \textbf{0.401} & 0.162          & 0.296          & \textbf{0.158} & \textbf{0.282} & 0.181           & 0.304           & \textbf{0.179} & \textbf{0.302} & 0.181          & 0.304          & \textbf{0.179} & \textbf{0.301} \\
\multicolumn{1}{c|}{}                             & 336 & 1.791         & 1.070        & \textbf{1.091} & \textbf{0.874} & 0.511          & \textbf{0.535} & \textbf{0.345} & 0.575          & 0.369          & 0.443          & \textbf{0.281} & \textbf{0.375} & 0.792          & 0.652          & \textbf{0.596} & \textbf{0.598} & 0.333          & 0.441          & \textbf{0.293} & \textbf{0.431} & 0.338           & 0.422           & \textbf{0.322} & \textbf{0.415} & 0.324          & 0.413          & \textbf{0.323} & \textbf{0.411} \\
\multicolumn{1}{c|}{}                             & 720 & 2.920         & 1.410        & \textbf{0.906} & \textbf{0.750} & 1.139          & 0.832          & \textbf{0.920} & \textbf{0.727} & 1.407          & 0.850          & \textbf{0.842} & \textbf{0.639} & 1.609          & 0.978          & \textbf{1.210} & \textbf{0.850} & 0.898          & \textbf{0.725} & \textbf{0.821} & 0.782          & \textbf{0.853}  & \textbf{0.696}  & 0.995          & 0.761          & \textbf{0.846} & \textbf{0.696} & 1.117          & 0.785          \\ \cline{2-30} 
\rowcolor{pink!50}
\multicolumn{1}{c|}{\cellcolor{white}}                             & Avg & 1.726         & 1.034        & \textbf{0.845} & \textbf{0.752} & 0.525          & 0.511          & \textbf{0.410} & \textbf{0.477} & 0.516          & 0.456          & \textbf{0.344} & \textbf{0.364} & 0.844          & 0.660          & \textbf{0.549} & \textbf{0.528} & 0.369          & 0.418          & \textbf{0.338} & \textbf{0.423} & \textbf{0.364}  & \textbf{0.407}  & 0.395          & 0.421          & \textbf{0.360} & \textbf{0.406} & 0.426          & 0.427          \\ \midrule
\multicolumn{1}{c|}{\multirow{5}{*}{\rotatebox{90}{ILI}}}         & 24  & 2.902         & 1.175        & \textbf{2.751} & \textbf{1.117} & 4.724          & 1.509          & \textbf{3.660} & \textbf{1.355} & 1.431          & 0.797          & \textbf{1.326} & \textbf{0.696} & 3.224          & 1.276          & \textbf{3.127} & \textbf{1.219} & 2.280          & 1.061          & \textbf{2.249} & \textbf{1.057} & 2.443           & 1.078           & \textbf{2.389} & \textbf{1.038} & 3.489          & 1.373          & \textbf{2.193} & \textbf{0.987} \\
\multicolumn{1}{c|}{}                             & 36  & 2.897         & 1.182        & \textbf{2.745} & \textbf{1.178} & 4.914          & 1.547          & \textbf{3.987} & \textbf{1.388} & 1.443          & 0.828          & \textbf{1.319} & \textbf{0.808} & 3.287          & 1.264          & \textbf{3.243} & \textbf{1.230} & 2.235          & 1.059          & \textbf{2.214} & \textbf{1.053} & 2.455           & 1.086           & \textbf{2.432} & \textbf{1.042} & 3.530          & 1.370          & \textbf{2.080} & \textbf{0.971} \\
\multicolumn{1}{c|}{}                             & 48  & 2.872         & 1.158        & \textbf{2.711} & \textbf{1.106} & 5.115          & 1.582          & \textbf{4.398} & \textbf{1.399} & 1.710          & 0.892          & \textbf{1.672} & \textbf{0.870} & 3.206          & 1.251          & \textbf{3.117} & \textbf{1.253} & 2.298          & 1.079          & \textbf{2.262} & \textbf{1.069} & 3.437           & 1.331           & \textbf{3.412} & \textbf{1.329} & 3.671          & 1.391          & \textbf{2.122} & \textbf{0.969} \\
\multicolumn{1}{c|}{}                             & 60  & 2.887         & 1.154        & \textbf{2.746} & \textbf{1.104} & 5.293          & 1.623          & \textbf{4.620} & \textbf{1.487} & 1.480          & 0.769          & \textbf{1.397} & \textbf{0.718} & \textbf{3.390} & \textbf{1.306} & 3.575          & 1.310          & 2.573          & 1.157          & \textbf{2.378} & \textbf{1.024} & 2.734           & 1.155           & \textbf{2.730} & \textbf{1.152} & 4.030          & 1.462          & \textbf{1.986} & \textbf{0.966} \\ \cline{2-30} 
\rowcolor{pink!50}
\multicolumn{1}{c|}{\cellcolor{white}}                             & Avg & 2.889         & 1.167        & \textbf{2.738} & \textbf{1.126} & 5.012          & 1.565          & \textbf{4.166} & \textbf{1.407} & 1.516          & 0.821          & \textbf{1.428} & \textbf{0.773} & 3.277          & 1.274          & \textbf{3.265} & \textbf{1.253} & 2.347          & 1.089          & \textbf{2.276} & \textbf{1.051} & 2.767           & 1.162           & \textbf{2.741} & \textbf{1.140} & 3.680          & 1.399          & \textbf{2.095} & \textbf{0.973} \\ \midrule
\multicolumn{1}{c|}{\multirow{5}{*}{\rotatebox{90}{Electricity}}} & 96  & 0.322         & 0.409        & \textbf{0.319} & \textbf{0.405} & 0.204          & 0.319          & \textbf{0.201} & \textbf{0.372} & 0.278          & 0.353          & \textbf{0.249} & \textbf{0.325} & 0.183          & 0.285          & \textbf{0.177} & \textbf{0.279} & 0.195          & 0.277          & \textbf{0.189} & \textbf{0.268} & 0.148           & 0.239           & \textbf{0.130} & \textbf{0.229} & 0.200          & \textbf{0.278} & \textbf{0.197} & 0.285          \\
\multicolumn{1}{c|}{}                             & 192 & 0.346         & 0.430        & \textbf{0.307} & \textbf{0.415} & 0.223          & 0.330          & \textbf{0.209} & \textbf{0.313} & 0.257          & 0.335          & \textbf{0.213} & \textbf{0.304} & 0.207          & 0.306          & \textbf{0.201} & \textbf{0.302} & \textbf{0.194} & 0.280          & 0.199          & \textbf{0.279} & 0.167           & 0.258           & \textbf{0.164} & \textbf{0.234} & 0.200          & 0.281          & \textbf{0.198} & \textbf{0.280} \\
\multicolumn{1}{c|}{}                             & 336 & 0.355         & 0.436        & \textbf{0.348} & \textbf{0.408} & 0.237          & 0.342          & \textbf{0.216} & \textbf{0.313} & 0.273          & 0.350          & \textbf{0.259} & \textbf{0.312} & 0.213          & 0.315          & \textbf{0.208} & \textbf{0.309} & 0.207          & 0.296          & \textbf{0.204} & \textbf{0.289} & 0.178           & 0.271           & \textbf{0.169} & \textbf{0.269} & 0.214          & 0.295          & \textbf{0.213} & \textbf{0.295} \\
\multicolumn{1}{c|}{}                             & 720 & 0.388         & 0.452        & \textbf{0.373} & \textbf{0.421} & 0.337          & 0.405          & \textbf{0.317} & \textbf{0.402} & 0.230          & 0.311          & \textbf{0.210} & \textbf{0.301} & 0.251          & 0.338          & \textbf{0.248} & \textbf{0.319} & 0.243          & 0.328          & \textbf{0.239} & \textbf{0.323} & 0.209           & 0.298           & \textbf{0.205} & \textbf{0.284} & 0.256          & 0.328          & \textbf{0.255} & \textbf{0.328} \\ \cline{2-30} 
\rowcolor{pink!50}
\multicolumn{1}{c|}{\cellcolor{white}}                             & Avg & 0.352         & 0.432        & \textbf{0.337} & \textbf{0.412} & 0.250          & \textbf{0.349} & \textbf{0.236} & 0.350          & 0.259          & 0.337          & \textbf{0.233} & \textbf{0.311} & 0.213          & 0.311          & \textbf{0.209} & \textbf{0.302} & 0.210          & 0.296          & \textbf{0.208} & \textbf{0.290} & 0.176           & 0.267           & \textbf{0.167} & \textbf{0.254} & 0.217          & \textbf{0.295} & \textbf{0.216} & 0.297          \\ \midrule
\multicolumn{1}{c|}{\multirow{5}{*}{\rotatebox{90}{Traffic}}}     & 96  & 0.742         & 0.414        & \textbf{0.726} & \textbf{0.405} & 0.645          & 0.414          & \textbf{0.505} & \textbf{0.417} & 0.446          & 0.283          & \textbf{0.403} & \textbf{0.215} & 0.630          & 0.365          & \textbf{0.503} & \textbf{0.274} & 0.650          & 0.397          & \textbf{0.630} & \textbf{0.371} & 0.392           & 0.268           & \textbf{0.383} & \textbf{0.262} & \textbf{0.658} & \textbf{0.409} & 0.687          & 0.422          \\
\multicolumn{1}{c|}{}                             & 192 & 0.759         & 0.428        & \textbf{0.739} & \textbf{0.416} & 0.619          & 0.387          & \textbf{0.507} & \textbf{0.304} & 0.453          & 0.286          & \textbf{0.408} & \textbf{0.252} & 0.541          & 0.334          & \textbf{0.473} & \textbf{0.249} & 0.600          & 0.372          & \textbf{0.573} & \textbf{0.347} & 0.413           & 0.277           & \textbf{0.411} & \textbf{0.273} & 0.620          & 0.371          & \textbf{0.592} & \textbf{0.358} \\
\multicolumn{1}{c|}{}                             & 336 & 0.877         & 0.494        & \textbf{0.872} & \textbf{0.429} & 0.621          & 0.380          & \textbf{0.557} & \textbf{0.368} & 0.468          & 0.291          & \textbf{0.429} & \textbf{0.270} & 0.592          & 0.365          & \textbf{0.583} & \textbf{0.339} & 0.606          & 0.374          & \textbf{0.584} & \textbf{0.350} & 0.425           & 0.283           & \textbf{0.420} & \textbf{0.279} & 0.619          & 0.368          & \textbf{0.586} & \textbf{0.346} \\
\multicolumn{1}{c|}{}                             & 720 & 1.034         & 0.581        & \textbf{0.934} & \textbf{0.523} & 0.718          & 0.442          & \textbf{0.639} & \textbf{0.385} & \textbf{0.594} & \textbf{0.384} & 0.598          & 0.389          & 0.684          & 0.426          & \textbf{0.548} & \textbf{0.386} & 0.646          & 0.396          & \textbf{0.602} & \textbf{0.382} & 0.458           & 0.300           & \textbf{0.449} & \textbf{0.296} & 0.669          & 0.391          & \textbf{0.632} & \textbf{0.372} \\ \cline{2-30} 
\rowcolor{pink!50}
\multicolumn{1}{c|}{\cellcolor{white}}                             & Avg & 0.853         & 0.479        & \textbf{0.818} & \textbf{0.443} & 0.651          & 0.406          & \textbf{0.552} & \textbf{0.369} & 0.490          & 0.311          & \textbf{0.460} & \textbf{0.282} & 0.612          & 0.373          & \textbf{0.527} & \textbf{0.312} & 0.625          & 0.385          & \textbf{0.597} & \textbf{0.363} & 0.422           & 0.282           & \textbf{0.416} & \textbf{0.278} & 0.642          & 0.385          & \textbf{0.624} & \textbf{0.375} \\ \midrule
\multicolumn{1}{c|}{\multirow{5}{*}{\rotatebox{90}{Solar Wind}}}  & 96  & 1.710         & 0.759        & \textbf{0.920} & \textbf{0.633} & 1.193          & 0.765          & \textbf{0.950} & \textbf{0.643} & 1.121          & 0.687          & \textbf{0.914} & \textbf{0.629} & 1.113          & 0.676          & \textbf{0.980} & \textbf{0.612} & 1.008          & 0.657          & \textbf{0.906} & \textbf{0.614} & 1.210           & 0.739           & \textbf{0.912} & \textbf{0.623} & 1.234          & 0.762          & \textbf{1.089} & \textbf{0.690} \\
\multicolumn{1}{c|}{}                             & 192 & 1.991         & 0.801        & \textbf{1.027} & \textbf{0.687} & 1.530          & 0.890          & \textbf{1.068} & \textbf{0.689} & 1.130          & 0.737          & \textbf{1.017} & \textbf{0.674} & 1.205          & 0.714          & \textbf{1.142} & \textbf{0.679} & 1.076          & 0.691          & \textbf{1.021} & \textbf{0.667} & 1.433           & 0.828           & \textbf{1.029} & \textbf{0.675} & 1.410          & 0.833          & \textbf{1.291} & \textbf{0.779} \\
\multicolumn{1}{c|}{}                             & 336 & 1.958         & 0.826        & \textbf{1.087} & \textbf{0.714} & 1.437          & 0.852          & \textbf{1.108} & \textbf{0.706} & 1.137          & 0.741          & \textbf{1.039} & \textbf{0.688} & 1.221          & 0.724          & \textbf{1.157} & \textbf{0.685} & 1.100          & 0.702          & \textbf{1.079} & \textbf{0.689} & 1.415           & 0.820           & \textbf{1.088} & \textbf{0.699} & 1.394          & 0.825          & \textbf{1.304} & \textbf{0.781} \\
\multicolumn{1}{c|}{}                             & 720 & 2.154         & 0.823        & \textbf{1.065} & \textbf{0.702} & 1.286          & 0.799          & \textbf{1.103} & \textbf{0.702} & 1.046          & 0.701          & \textbf{0.821} & \textbf{0.673} & 1.155          & 0.699          & \textbf{1.086} & \textbf{0.661} & 1.097          & 0.700          & \textbf{1.072} & \textbf{0.688} & 1.381           & 0.804           & \textbf{1.083} & \textbf{0.703} & 1.358          & 0.814          & \textbf{1.275} & \textbf{0.773} \\ \cline{2-30} 
\rowcolor{pink!50}
\multicolumn{1}{c|}{\cellcolor{white}}                             & Avg & 1.953         & 0.803        & \textbf{1.025} & \textbf{0.684} & 1.362          & 0.826          & \textbf{1.057} & \textbf{0.685} & 1.109          & 0.717          & \textbf{0.948} & \textbf{0.666} & 1.174          & 0.703          & \textbf{1.091} & \textbf{0.659} & 1.071          & 0.687          & \textbf{1.019} & \textbf{0.665} & 1.360           & 0.798           & \textbf{1.028} & \textbf{0.675} & 1.349          & 0.809          & \textbf{1.239} & \textbf{0.756} \\ \bottomrule

\end{tabular}}
\caption{Multivariate long sequence time-series forecasting results. We report the MSE/MAE of different prediction lengths. The look-up window is set to $L = 336$ for PatchTST, DLinear, and SCINet, and $L = 96$ for other models. The \textbf{best results} are highlighted in \textbf{bold}.}
\label{tab:ALL SOTA Multivariate Quantitative Results}
\end{table*}

\begin{table*}[!t]
\centering
\centering\setlength{\tabcolsep}{3pt} 
\renewcommand{\arraystretch}{1.5} 
\resizebox{1.0\textwidth}{!}{
\large
\begin{tabular}{cc|cccc|cccc|cccc|cccc|cccc|cccc|cccc}
\toprule
\multicolumn{2}{c|}{Model}        & \multicolumn{2}{c}{Informer} & \multicolumn{2}{c|}{+HCAN}      & \multicolumn{2}{c}{Autoformer} & \multicolumn{2}{c|}{+HCAN}      & \multicolumn{2}{c}{PatchTST}    & \multicolumn{2}{c|}{+HCAN}      & \multicolumn{2}{c}{SCINet}      & \multicolumn{2}{c|}{+HCAN}      & \multicolumn{2}{c}{Dlinear}     & \multicolumn{2}{c|}{+HCAN}      & \multicolumn{2}{c}{iTransforrmer} & \multicolumn{2}{c|}{+HCAN}      & \multicolumn{2}{c}{FITS} & \multicolumn{2}{c}{+HCAN}     \\
\multicolumn{2}{c|}{Metric}       & MSE           & MAE          & MSE            & MAE            & MSE            & MAE           & MSE            & MAE            & MSE            & MAE            & MSE            & MAE            & MSE            & MAE            & MSE            & MAE            & MSE            & MAE            & MSE            & MAE            & MSE             & MAE             & MSE            & MAE            & MSE         & MAE        & MSE            & MAE          \\ \midrule
\multirow{4}{*}{\rotatebox{90}{ETTh1}}      & 96  & 0.255         & 0.438        & \textbf{0.121} & \textbf{0.283} & 0.088          & 0.234         & \textbf{0.082} & \textbf{0.223} & 0.055          & \textbf{0.179} & \textbf{0.055} & 0.181          & 0.088          & 0.227          & \textbf{0.068} & \textbf{0.197} & 0.057          & 0.179          & \textbf{0.053} & \textbf{0.178} & 0.061           & 0.190           & \textbf{0.060} & \textbf{0.187} & 0.056       & 0.179      & \textbf{0.054} & \textbf{0.178}   \\
                            & 192 & 0.283         & 0.461        & \textbf{0.092} & \textbf{0.236} & 0.108          & 0.252         & \textbf{0.086} & \textbf{0.223} & \textbf{0.071} & \textbf{0.205} & 0.072          & 0.206          & 0.105          & 0.249          & \textbf{0.084} & \textbf{0.221} & 0.077          & 0.210          & \textbf{0.075} & \textbf{0.209} & 0.073           & 0.206           & \textbf{0.072} & \textbf{0.205} & 0.075       & 0.210      & \textbf{0.072} & \textbf{0.209}  \\
                            & 336 & 0.291         & 0.469        & \textbf{0.088} & \textbf{0.230} & 0.118          & 0.268         & \textbf{0.091} & \textbf{0.237} & 0.082          & 0.227          & \textbf{0.078} & \textbf{0.217} & 0.130          & 0.286          & \textbf{0.094} & \textbf{0.244} & 0.097          & 0.244          & \textbf{0.088} & \textbf{0.235} & 0.089           & 0.231           & \textbf{0.087} & \textbf{0.230} & 0.091       & 0.237      & \textbf{0.089} & \textbf{0.236}   \\
                            & 720 & 0.256         & 0.426        & \textbf{0.106} & \textbf{0.260} & 0.138          & 0.298         & \textbf{0.121} & \textbf{0.279} & 0.086          & 0.232          & \textbf{0.081} & \textbf{0.204} & 0.214          & 0.387          & \textbf{0.134} & \textbf{0.292} & 0.168          & 0.336          & \textbf{0.164} & \textbf{0.331} & \textbf{0.083}  & \textbf{0.226}  & 0.105          & 0.258          & 0.104       & 0.254      & \textbf{0.096} & \textbf{0.245}   \\ \midrule
\multirow{4}{*}{\rotatebox{90}{ETTh2}}      & 96  & 0.302         & 0.446        & \textbf{0.182} & \textbf{0.349} & 0.169          & 0.321         & \textbf{0.140} & \textbf{0.295} & 0.129          & 0.282          & \textbf{0.127} & \textbf{0.278} & 0.130          & 0.281          & \textbf{0.129} & \textbf{0.280} & 0.133          & 0.281          & \textbf{0.128} & \textbf{0.271} & 0.135           & 0.286           & \textbf{0.133} & \textbf{0.283} & 0.125       & 0.269      & \textbf{0.123} & \textbf{0.268}  \\
                            & 192 & 0.264         & 0.414        & \textbf{0.206} & \textbf{0.365} & 0.211          & 0.359         & \textbf{0.179} & \textbf{0.328} & 0.169          & 0.328          & \textbf{0.162} & \textbf{0.305} & 0.327          & 0.459          & \textbf{0.169} & \textbf{0.326} & 0.177          & 0.330          & \textbf{0.174} & \textbf{0.325} & 0.182           & 0.336           & \textbf{0.178} & \textbf{0.334} & 0.177       & 0.327      & \textbf{0.174} & \textbf{0.325} \\
                            & 336 & 0.324         & 0.456        & \textbf{0.223} & \textbf{0.385} & 0.255          & 0.398         & \textbf{0.226} & \textbf{0.373} & 0.187          & 0.352          & \textbf{0.187} & \textbf{0.340} & \textbf{0.198} & \textbf{0.358} & 0.220          & 0.378          & \textbf{0.212} & \textbf{0.369} & 0.225          & 0.375          & 0.218           & 0.373           & \textbf{0.215} & \textbf{0.371} & 0.222       & 0.375      & \textbf{0.221} & \textbf{0.371}   \\
                            & 720 & 0.302         & 0.447        & \textbf{0.249} & \textbf{0.408} & 0.334          & 0.459         & \textbf{0.292} & \textbf{0.432} & 0.224          & 0.383          & \textbf{0.201} & \textbf{0.357} & 0.486          & 0.569          & \textbf{0.221} & \textbf{0.377} & 0.298          & 0.444          & \textbf{0.259} & \textbf{0.413} & 0.240           & 0.391           & \textbf{0.238} & \textbf{0.389} & 0.258       & 0.409      & \textbf{0.255} & \textbf{0.406}  \\ \midrule
\multirow{4}{*}{\rotatebox{90}{ETTm1}}      & 96  & 0.093         & 0.249        & \textbf{0.046} & \textbf{0.166} & 0.059          & 0.186         & \textbf{0.047} & \textbf{0.167} & 0.026          & \textbf{0.121} & \textbf{0.024} & 0.123          & 0.049          & 0.170          & \textbf{0.029} & \textbf{0.127} & 0.030          & 0.128          & \textbf{0.026} & \textbf{0.125} & 0.029           & 0.128           & \textbf{0.028} & \textbf{0.124} & 0.029       & 0.127      & \textbf{0.027} & \textbf{0.126}  \\
                            & 192 & 0.232         & 0.404        & \textbf{0.059} & \textbf{0.189} & 0.081          & 0.223         & \textbf{0.057} & \textbf{0.187} & 0.039          & 0.150          & \textbf{0.037} & \textbf{0.148} & 0.077          & 0.215          & \textbf{0.049} & \textbf{0.166} & 0.044          & 0.155          & \textbf{0.043} & \textbf{0.151} & 0.049           & 0.169           & \textbf{0.045} & \textbf{0.167} & 0.043       & 0.158      & \textbf{0.042} & \textbf{0.155}  \\
                            & 336 & 0.271         & 0.453        & \textbf{0.108} & \textbf{0.264} & 0.088          & 0.242         & \textbf{0.072} & \textbf{0.205} & 0.053          & 0.173          & \textbf{0.050} & \textbf{0.168} & 0.109          & 0.259          & \textbf{0.089} & \textbf{0.229} & 0.064          & 0.187          & \textbf{0.059} & \textbf{0.183} & 0.061           & 0.190           & \textbf{0.060} & \textbf{0.187} & 0.057       & 0.183      & \textbf{0.056} & \textbf{0.181}  \\
                            & 720 & 0.464         & 0.606        & \textbf{0.118} & \textbf{0.277} & 0.122          & 0.275         & \textbf{0.079} & \textbf{0.214} & 0.074          & 0.207          & \textbf{0.070} & \textbf{0.203} & 0.139          & 0.296          & \textbf{0.117} & \textbf{0.261} & \textbf{0.081} & \textbf{0.211} & 0.082          & 0.216          & 0.083           & 0.220           & \textbf{0.082} & \textbf{0.218} & 0.079       & 0.216      & \textbf{0.075} & \textbf{0.216}  \\ \midrule
\multirow{4}{*}{\rotatebox{90}{ETTm2}}      & 96  & 0.092         & 0.233        & \textbf{0.065} & \textbf{0.209} & 0.127          & 0.274         & \textbf{0.095} & \textbf{0.239} & 0.065          & 0.186          & \textbf{0.065} & \textbf{0.185} & 0.079          & 0.216          & \textbf{0.069} & \textbf{0.195} & 0.064          & 0.184          & \textbf{0.061} & \textbf{0.181} & 0.069           & 0.189           & \textbf{0.069} & \textbf{0.187} & 0.070       & 0.190      & \textbf{0.069} & \textbf{0.189}   \\
                            & 192 & 0.134         & 0.283        & \textbf{0.107} & \textbf{0.255} & 0.146          & 0.295         & \textbf{0.123} & \textbf{0.270} & 0.094          & 0.231          & \textbf{0.091} & \textbf{0.227} & 0.105          & 0.252          & \textbf{0.094} & \textbf{0.232} & 0.092          & 0.227          & \textbf{0.087} & \textbf{0.217} & 0.107           & 0.244           & \textbf{0.106} & \textbf{0.242} & 0.100       & 0.235      & \textbf{0.098} & \textbf{0.233}   \\
                            & 336 & 0.178         & 0.340        & \textbf{0.141} & \textbf{0.298} & 0.217          & 0.359         & \textbf{0.126} & \textbf{0.278} & 0.120          & 0.265          & \textbf{0.117} & \textbf{0.259} & 0.130          & 0.282          & \textbf{0.128} & \textbf{0.276} & 0.129          & 0.273          & \textbf{0.120} & \textbf{0.262} & 0.144           & 0.289           & \textbf{0.143} & \textbf{0.286} & 0.128       & 0.271      & \textbf{0.126} & \textbf{0.270}   \\
                            & 720 & 0.221         & 0.375        & \textbf{0.156} & \textbf{0.313} & 0.198          & 0.348         & \textbf{0.184} & \textbf{0.335} & 0.172          & 0.322          & \textbf{0.169} & \textbf{0.310} & 0.175          & 0.328          & \textbf{0.155} & \textbf{0.307} & \textbf{0.176} & \textbf{0.321} & 0.181          & 0.326          & \textbf{0.185}  & 0.337           & 0.187          & \textbf{0.334} & 0.178       & 0.326      & \textbf{0.176} & \textbf{0.324}   \\ \midrule
\multirow{4}{*}{\rotatebox{90}{Solar Wind}} & 96  & 1.443         & 0.892        & \textbf{1.268} & \textbf{0.823} & 2.316          & 1.220         & \textbf{1.289} & \textbf{0.870} & 1.021          & 0.687          & \textbf{0.851} & \textbf{0.663} & 1.518          & 0.885          & \textbf{1.366} & \textbf{0.815} & 1.316          & 0.849          & \textbf{1.223} & \textbf{0.812} & 1.727           & 0.977           & \textbf{1.266} & \textbf{0.823} & 1.669       & 0.969      & \textbf{1.658} & \textbf{0.954} \\
                            & 192 & 1.765         & 1.003        & \textbf{1.581} & \textbf{0.963} & 2.765          & 1.364         & \textbf{1.590} & \textbf{0.965} & 1.130          & 0.757          & \textbf{1.030} & \textbf{0.738} & 1.836          & 1.003          & \textbf{1.723} & \textbf{0.952} & 1.568          & 0.941          & \textbf{1.549} & \textbf{0.934} & 2.273           & 1.179           & \textbf{1.568} & \textbf{0.948} & 2.308       & 1.198      & \textbf{2.280} & \textbf{1.174}  \\
                            & 336 & 1.849         & 1.047        & \textbf{1.740} & \textbf{1.023} & 2.783          & 1.351         & \textbf{1.715} & \textbf{1.013} & 1.137          & 0.791          & \textbf{1.098} & \textbf{0.747} & 1.853          & 1.020          & \textbf{1.746} & \textbf{0.979} & 1.686          & 0.998          & \textbf{1.671} & \textbf{0.995} & 2.370           & 1.218           & \textbf{1.714} & \textbf{1.010} & 2.355       & 1.220      & \textbf{2.327} & \textbf{1.200}  \\
                            & 720 & 1.826         & 1.052        & \textbf{1.694} & \textbf{1.019} & 2.606          & 1.300         & \textbf{1.701} & \textbf{1.022} & 1.125          & 0.792          & \textbf{1.041} & \textbf{0.703} & 1.672          & 0.976          & \textbf{1.547} & \textbf{0.933} & 1.660          & 0.997          & \textbf{1.654} & \textbf{0.990} & 2.228           & 1.183           & \textbf{1.679} & \textbf{1.001} & 2.220       & 1.185      & \textbf{2.189} & \textbf{1.163}  \\ \bottomrule
\end{tabular}}
\caption{Univariate long sequence time-series forecasting results on ETT full benchmark and Solar Wind dataset. We report the MSE/MAE of different prediction lengths $T \in \{96, 192, 336, 720\}$ for comparison. The look-up window is set to $L = 336$ for PatchTST, DLinear, and SCINet, and $L = 96$ for other models. The \textbf{best results} are highlighted in \textbf{bold}.}
\label{tab:ALL SOTA Univariate Quantitative Results}
\end{table*}

\begin{table*}[!t]
\centering
\centering\setlength{\tabcolsep}{8pt} 
\renewcommand{\arraystretch}{1.0} 
\resizebox{0.8\textwidth}{!}{
\begin{tabular}{cc|cccccc|cccccc}
\toprule
\multicolumn{2}{c|}{\multirow{2}{*}{Model}}       & \multicolumn{6}{c|}{PatchTST}                                                                                                                        & \multicolumn{6}{c}{FITS}                                                                                                                            \\ \cline{3-14} 
\multicolumn{2}{c|}{}                             & \multicolumn{2}{c|}{+ HCAN}                          & \multicolumn{2}{c|}{+ MAE}                           & \multicolumn{2}{c|}{+ Ordinal Entropy} & \multicolumn{2}{c|}{+ HCAN}                          & \multicolumn{2}{c|}{+ MAE}                           & \multicolumn{2}{c}{+ Ordinal Entropy} \\
\multicolumn{2}{c|}{Metric}                       & MSE            & \multicolumn{1}{c|}{MAE}            & MSE            & \multicolumn{1}{c|}{MAE}            & MSE                & MAE               & MSE            & \multicolumn{1}{c|}{MAE}            & MSE            & \multicolumn{1}{c|}{MAE}            & MSE               & MAE               \\ \midrule
\multicolumn{1}{c|}{\multirow{4}{*}{ETTh1}} & 96  & \textbf{0.358} & \multicolumn{1}{c|}{0.398}          & 0.367          & \multicolumn{1}{c|}{\textbf{0.392}} & 0.389              & 0.394             & \textbf{0.377} & \multicolumn{1}{c|}{\textbf{0.388}} & 0.384          & \multicolumn{1}{c|}{0.388}          & 0.398             & 0.423             \\
\multicolumn{1}{c|}{}                       & 192 & \textbf{0.382} & \multicolumn{1}{c|}{\textbf{0.412}} & 0.411          & \multicolumn{1}{c|}{0.416}          & 0.445              & 0.456             & \textbf{0.429} & \multicolumn{1}{c|}{\textbf{0.417}} & 0.436          & \multicolumn{1}{c|}{0.418}          & 0.469             & 0.479             \\
\multicolumn{1}{c|}{}                       & 336 & \textbf{0.421} & \multicolumn{1}{c|}{0.432}          & 0.431          & \multicolumn{1}{c|}{\textbf{0.427}} & 0.452              & 0.472             & \textbf{0.481} & \multicolumn{1}{c|}{\textbf{0.436}} & 0.478          & \multicolumn{1}{c|}{0.439}          & 0.498             & 0.521             \\
\multicolumn{1}{c|}{}                       & 720 & \textbf{0.425} & \multicolumn{1}{c|}{\textbf{0.451}} & 0.439          & \multicolumn{1}{c|}{0.455}          & 0.453              & 0.461             & \textbf{0.456} & \multicolumn{1}{c|}{0.475}          & 0.462          & \multicolumn{1}{c|}{\textbf{0.455}} & 0.472             & 0.489             \\ \midrule
\multicolumn{1}{c|}{\multirow{4}{*}{ETTh2}} & 96  & \textbf{0.265} & \multicolumn{1}{c|}{0.334}          & 0.277          & \multicolumn{1}{c|}{\textbf{0.331}} & 0.323              & 0.347             & \textbf{0.284} & \multicolumn{1}{c|}{\textbf{0.322}} & 0.292          & \multicolumn{1}{c|}{0.337}          & 0.334             & 0.348             \\
\multicolumn{1}{c|}{}                       & 192 & \textbf{0.323} & \multicolumn{1}{c|}{\textbf{0.363}} & 0.346          & \multicolumn{1}{c|}{0.377}          & 0.356              & 0.372             & \textbf{0.372} & \multicolumn{1}{c|}{\textbf{0.382}} & 0.377          & \multicolumn{1}{c|}{0.389}          & 0.401             & 0.417             \\
\multicolumn{1}{c|}{}                       & 336 & \textbf{0.368} & \multicolumn{1}{c|}{0.401}          & 0.372          & \multicolumn{1}{c|}{\textbf{0.378}} & 0.397              & 0.413             & \textbf{0.408} & \multicolumn{1}{c|}{\textbf{0.419}} & 0.419          & \multicolumn{1}{c|}{0.425}          & 0.438             & 0.446             \\
\multicolumn{1}{c|}{}                       & 720 & 0.416          & \multicolumn{1}{c|}{0.440}          & \textbf{0.385} & \multicolumn{1}{c|}{\textbf{0.416}} & 0.439              & 0.456             & \textbf{0.409} & \multicolumn{1}{c|}{\textbf{0.421}} & 0.419          & \multicolumn{1}{c|}{0.436}          & 0.437             & 0.446             \\ \midrule
\multicolumn{1}{c|}{\multirow{4}{*}{ETTm1}} & 96  & \textbf{0.281} & \multicolumn{1}{c|}{\textbf{0.329}} & 0.293          & \multicolumn{1}{c|}{0.329}          & 0.293              & 0.302             & \textbf{0.336} & \multicolumn{1}{c|}{0.370}          & 0.337          & \multicolumn{1}{c|}{\textbf{0.353}} & 0.392             & 0.402             \\
\multicolumn{1}{c|}{}                       & 192 & \textbf{0.318} & \multicolumn{1}{c|}{\textbf{0.342}} & 0.337          & \multicolumn{1}{c|}{0.360}          & 0.351              & 0.363             & \textbf{0.383} & \multicolumn{1}{c|}{0.395}          & 0.385          & \multicolumn{1}{c|}{\textbf{0.376}} & 0.458             & 0.469             \\
\multicolumn{1}{c|}{}                       & 336 & \textbf{0.349} & \multicolumn{1}{c|}{\textbf{0.351}} & 0.381          & \multicolumn{1}{c|}{0.386}          & 0.372              & 0.384             & \textbf{0.408} & \multicolumn{1}{c|}{0.410}          & 0.418          & \multicolumn{1}{c|}{\textbf{0.398}} & 0.469             & 0.483             \\
\multicolumn{1}{c|}{}                       & 720 & 0.452          & \multicolumn{1}{c|}{0.432}          & \textbf{0.431} & \multicolumn{1}{c|}{\textbf{0.416}} & 0.473              & 0.483             & 0.492          & \multicolumn{1}{c|}{0.454}          & \textbf{0.486} & \multicolumn{1}{c|}{\textbf{0.436}} & 0.572             & 0.593             \\ \midrule
\multicolumn{1}{c|}{\multirow{4}{*}{ETTm2}} & 96  & \textbf{0.161} & \multicolumn{1}{c|}{\textbf{0.242}} & 0.162          & \multicolumn{1}{c|}{0.246}          & 0.169              & 0.173             & 0.181          & \multicolumn{1}{c|}{0.260}          & \textbf{0.183} & \multicolumn{1}{c|}{\textbf{0.258}} & 0.274             & 0.289             \\
\multicolumn{1}{c|}{}                       & 192 & \textbf{0.218} & \multicolumn{1}{c|}{0.293}          & 0.219          & \multicolumn{1}{c|}{\textbf{0.286}} & 0.253              & 0.265             & \textbf{0.231} & \multicolumn{1}{c|}{\textbf{0.290}} & 0.247          & \multicolumn{1}{c|}{0.299}          & 0.321             & 0.342             \\
\multicolumn{1}{c|}{}                       & 336 & \textbf{0.276} & \multicolumn{1}{c|}{0.330}          & 0.272          & \multicolumn{1}{c|}{\textbf{0.321}} & 0.352              & 0.361             & \textbf{0.306} & \multicolumn{1}{c|}{\textbf{0.335}} & 0.308          & \multicolumn{1}{c|}{0.338}          & 0.389             & 0.397             \\
\multicolumn{1}{c|}{}                       & 720 & \textbf{0.345} & \multicolumn{1}{c|}{0.382}          & 0.355          & \multicolumn{1}{c|}{\textbf{0.374}} & 0.398              & 0.413             & \textbf{0.403} & \multicolumn{1}{c|}{\textbf{0.330}} & 0.408          & \multicolumn{1}{c|}{0.394}          & 0.504             & 0.518             \\ \bottomrule
\end{tabular}}
\caption{Comparison with the regularization techniques.}
\label{tab:Regularization}
\end{table*}

\section{Full Forecasting Results}
The full multivariate forecasting results are provided in the following section due to the space limitation of the main text. Table~\ref{tab:ALL SOTA Multivariate Quantitative Results} presents the detailed multivariate results of all prediction lengths in terms of MSE/MAE across ten well-acknowledged benchmarks. And Table~\ref{tab:ALL SOTA Univariate Quantitative Results} provides the univariate results for MSE/MAE. Our proposed model consistently achieves state-of-the-art performance in real-world forecasting applications.

\section{Comparison with the regularization techniques}

\begin{table*}[!t]
\centering
\centering\setlength{\tabcolsep}{3pt} 
\renewcommand{\arraystretch}{1.0} 
\resizebox{0.6\textwidth}{!}{
\begin{tabular}{ccccc|cccc|cccc}
\toprule
\multicolumn{5}{c|}{Component}                                                                        & \multicolumn{4}{c|}{iTransformer}                                                        & \multicolumn{4}{c}{Dlinear}                                                              \\ \midrule
\multicolumn{2}{c}{UAC} & \multirow{2}{*}{Hierarchy} & \multirow{2}{*}{$\mathcal{L}_{HCL}$} & \multirow{2}{*}{HAA} & \multirow{2}{*}{96} & \multirow{2}{*}{192} & \multirow{2}{*}{336} & \multirow{2}{*}{720} & \multirow{2}{*}{96} & \multirow{2}{*}{192} & \multirow{2}{*}{336} & \multirow{2}{*}{720} \\ \cline{1-2}
$\mathcal{L}_{UAC}$     & $\mathcal{L}_{REG}$     &                            &                         &                      &                     &                      &                      &                      &                     &                      &                      &                    \\ \midrule
-          & -          & -                          & -                       & -                    & 0.176               & 0.225                & 0.281                & 0.358                & 0.175               & 0.218                & 0.263                & 0.332                \\
\checkmark          & -          & -                          & -                       & -                    & 0.173               & 0.225                & 0.285                & 0.354                & 0.171               & 0.215                & 0.261                & 0.337                \\
\checkmark          & \checkmark          & -                          & -                       & -                    & 0.174               & 0.223                & 0.278                & 0.350                & 0.173               & 0.211                & 0.262                & 0.334                \\
\checkmark          & \checkmark          & \checkmark                          & -                       & -                    & 0.168               & 0.220                & 0.277                & 0.343                & 0.170               & 0.215                & 0.265                & 0.324                \\
\checkmark          & \checkmark          & \checkmark                          & \checkmark                       & -                    & 0.167               & 0.221                & 0.276                & 0.341                & 0.169               & 0.207                & 0.262                & 0.320                \\
\checkmark          & \checkmark          & \checkmark                          & \checkmark                       & \checkmark                    & \textbf{0.161}      & \textbf{0.218}       & \textbf{0.276}       & \textbf{0.345}       & \textbf{0.164}      & \textbf{0.205}       & \textbf{0.258}       & \textbf{0.319}       \\ \bottomrule
\end{tabular}}
\caption{Ablation study of the components of HCAN on the Weather dataset using iTransformer and Dlinear as backbones.}
\label{tab:Ablation-appendix}
\end{table*}

\subsection{Mean Absolute Error Loss}
Traditional regularization methods, such as L1 or L2 penalties, primarily constrain model complexity to prevent overfitting. However, they may not effectively address the specific issue of over-smoothing in time series forecasting, where models fail to capture high-entropy features due to the limitations of Mean Squared Error (MSE) loss.
HCAN introduces a novel approach by reformulating the forecasting task as a classification problem, utilizing cross-entropy loss to better capture high-entropy features. This hierarchical structure enables the model to learn multi-granularity features, enhancing its ability to represent complex patterns in the data.

We replace the MSE with the MAE loss for PatchTST and iTransformer and report the results in Table~\ref{tab:Regularization}.
The key difference lies in how each loss function influences the model’s predictions. HCAN incorporates a cross-entropy term to promote diverse and informative features, enhancing model's ability to capture complex patterns. In contrast, MAE loss mainly focuses on reducing error but does not encourage feature diversity.

\subsection{High-entropy Loss}
In addition, we compared HCAN with Ordinal Entropy loss \cite{zhang2023improving} in Table~\ref{tab:Regularization}, which discretizes the continuous labels and treats each bin as a class to encourage higher-entropy feature spaces. We discretized the time series and replaced the MSE loss with Ordinal Entropy loss, conducting experiments on PatchTST and FITS. Clearly, our proposed HCAN achieves better performance, which further demonstrates the effectiveness of our approach.

\section{Ablation Study}

We conduct more ablation studies on the Weather dataset with iTransformer and DLinear further demonstrate the necessity of each model component, as shown in Table~\ref{tab:Ablation-appendix}.

Each component is essential to address specific challenges in time series forecasting:
\begin{itemize}
    \item Multi-Resolution Analysis: Captures patterns at different scales, essential for understanding both short-term fluctuations and long-term trends.
    \item Regularized Classification Loss: Mitigates over-confidence in predictions and enhances the model's generalization.
    \item Hierarchical Structure: Facilitates learning at various levels of abstraction, improving the model's capacity to handle complex temporal dependencies.
\end{itemize}

\section{Hyperparameter Sensitivity}
As shown in Table~\ref{tab:Hyperparameter-Sensitivity}, we conduct hyperparameter sensitivity experiments on $(K_c, K_f)$ based on Informer and achieve optimal performance with the configuration (2, 4), demonstrating that HCAN is robust to these hyperparameters and maintains consistent generalizability across diverse datasets.

\begin{table}[t]
\centering
\scalebox{0.7}{
\begin{tabular}{c|cc|cc|cc|cc}
\toprule
\multirow{2}{*}{Informer} & \multicolumn{2}{c|}{(2, 4)}     & \multicolumn{2}{c|}{(2, 8)} & \multicolumn{2}{c|}{(2, 16)} & \multicolumn{2}{c}{(2, 32)} \\
                          & MSE            & MAE            & MSE          & MAE          & MSE           & MAE          & MSE          & MAE          \\ \midrule
ETTh1                     & \textbf{0.703} & \textbf{0.617} & 0.791        & 0.668        & 0.783         & 0.665        & 0.834        & 0.700        \\
ETTh2                     & \textbf{1.563} & \textbf{1.011} & 1.568        & 1.015        & 1.844         & 1.135        & 1.774        & 1.109        \\
ETTm1                     & \textbf{0.592} & \textbf{0.544} & 0.596        & 0.628        & 0.628         & 0.559        & 0.614        & 0.579        \\
ETTm2                     & \textbf{0.337} & \textbf{0.445} & 0.369        & 0.459        & 0.400         & 0.488        & 0.419        & 0.505        \\ \bottomrule
\end{tabular}}
\caption{Hyperparameter sensitively experiments on $(K_c, K_f)$.}
\label{tab:Hyperparameter-Sensitivity}
\end{table}

\section{Exploring Multiple Hierarchical Levels}
The cited works employ multiple hierarchical levels to capture time-series patterns at various granularities. Incorporating more than three levels may enhance our model's ability to represent complex temporal structures. To explore this, we have conducted some experiments based on DLinear using both ETTh1 and ETTh2:

\begin{itemize}
    \item H=1 ($K_1=1$) (Backone)
    \item H=2 ($K_1=1$, $K_2=2$)
    \item H=3 ($K_1=1$, $K_2=2$, $K_3=4$) (HCAN)
    \item H=4 ($K_1=1$, $K_2=2$, $K_3=4$, $K_4=8$)
    \item H=5 ($K_1=1$, $K_2=2$, $K_3=4$, $K_4=8$, $K_5=16$)
\end{itemize}
where H denotes the number of hierarchy levels, and $K_i$ represents the number of classes. We report the results in Table~\ref{tab:Hierarchical Levels}. 
The best performance is achieved with H=3 in HCAN. While incorporating more levels of granularity could provide benefits compared to H=1, our experiments suggest that three levels are sufficient to capture the necessary temporal dependencies without introducing excessive complexity.

\begin{table}[t]
\centering
\scalebox{0.7}{
\begin{tabular}{c|cccc|cccc}
\toprule
\multirow{2}{*}{\begin{tabular}[c]{@{}c@{}}Number of\\ hierarchies\end{tabular}} & \multicolumn{4}{c|}{ETTh1}                                        & \multicolumn{4}{c}{ETTh2}                                         \\
                                                                                 & 96             & 192            & 336            & 720            & 96             & 192            & 336            & 720            \\ \midrule
H=1                                                                              & 0.384          & 0.443          & 0.451          & 0.535          & 0.300          & 0.394          & 0.465          & 0.733          \\
H=2                                                                              & 0.396          & 0.405          & 0.457          & 0.524          & 0.294          & 0.369          & 0.452          & 0.583          \\
H=3                                                                              & \textbf{0.371} & \textbf{0.420} & \textbf{0.439} & \textbf{0.482} & \textbf{0.287} & \textbf{0.359} & \textbf{0.439} & \textbf{0.557} \\
H=4                                                                              & 0.397          & 0.430          & 0.451          & 0.559          & 0.291          & 0.379          & 0.458          & 0.589          \\
H=5                                                                              & 0.370          & 0.421          & 0.475          & 0.530          & 0.303          & 0.384          & 0.461          & 0.593          \\ \bottomrule
\end{tabular}}
\caption{Experiments on the hierarchical levels.}
\label{tab:Hierarchical Levels}
\end{table}

\begin{figure*}[!t]
	\centering
	\begin{subfigure}[t]{0.20\textwidth} 
		\includegraphics[width=\textwidth]{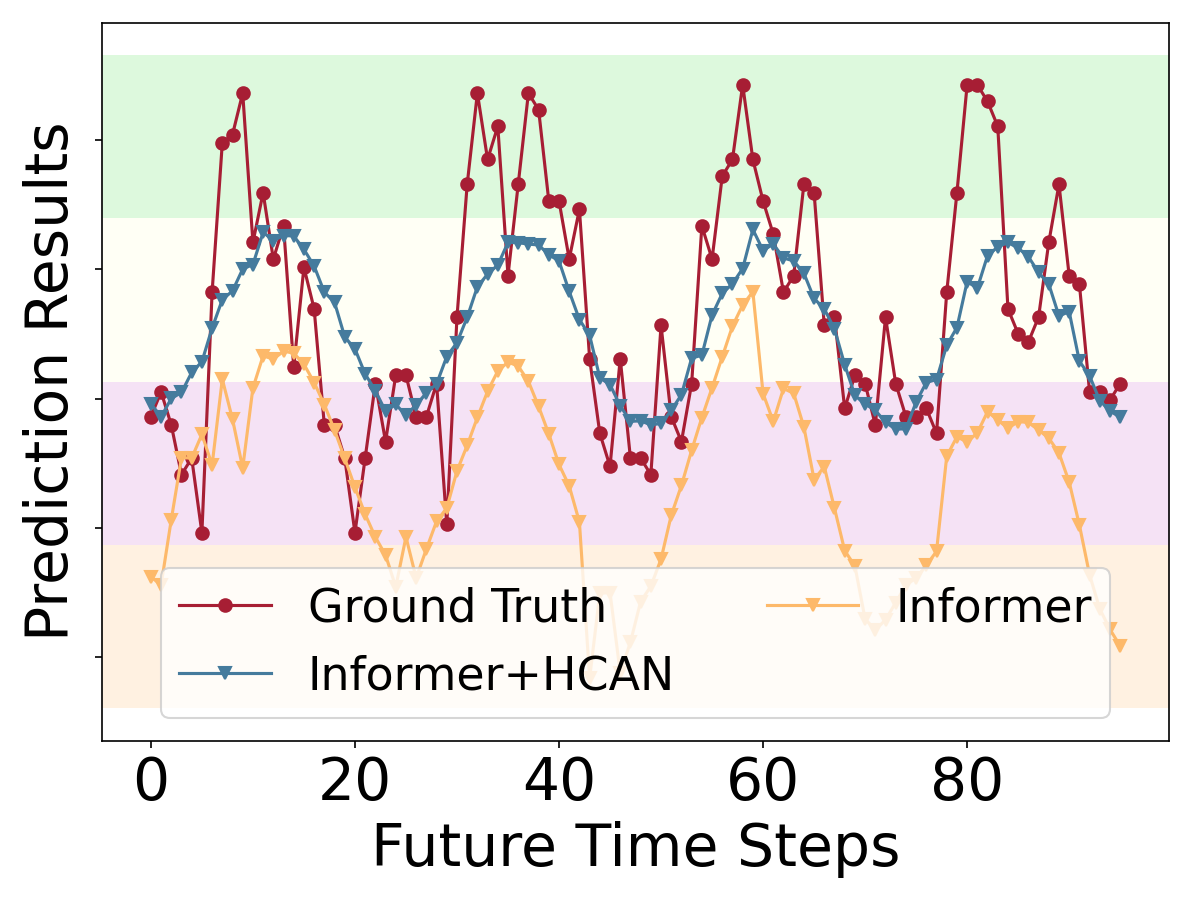}
		\caption{Informer+HCAN}
		\label{fig:Multivariate-Informer-96}
	\end{subfigure}
	\hspace{1cm}
	\begin{subfigure}[t]{0.20\textwidth}
		\includegraphics[width=\textwidth]{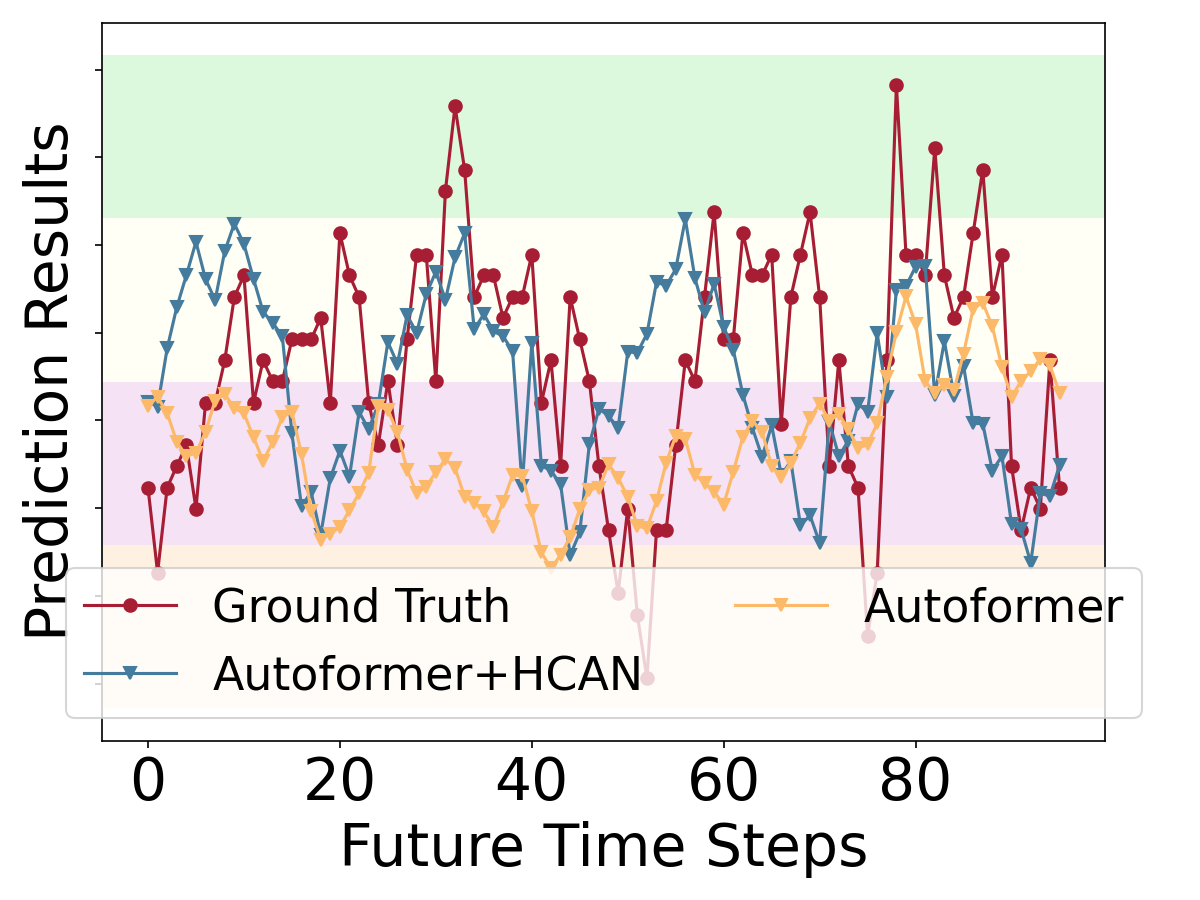}
		\caption{Autoformer+HCAN}
		\label{fig:Multivariate-Autoformer-96}
	\end{subfigure}
        \hspace{1cm}
	\begin{subfigure}[t]{0.20\textwidth}
		\includegraphics[width=\textwidth]{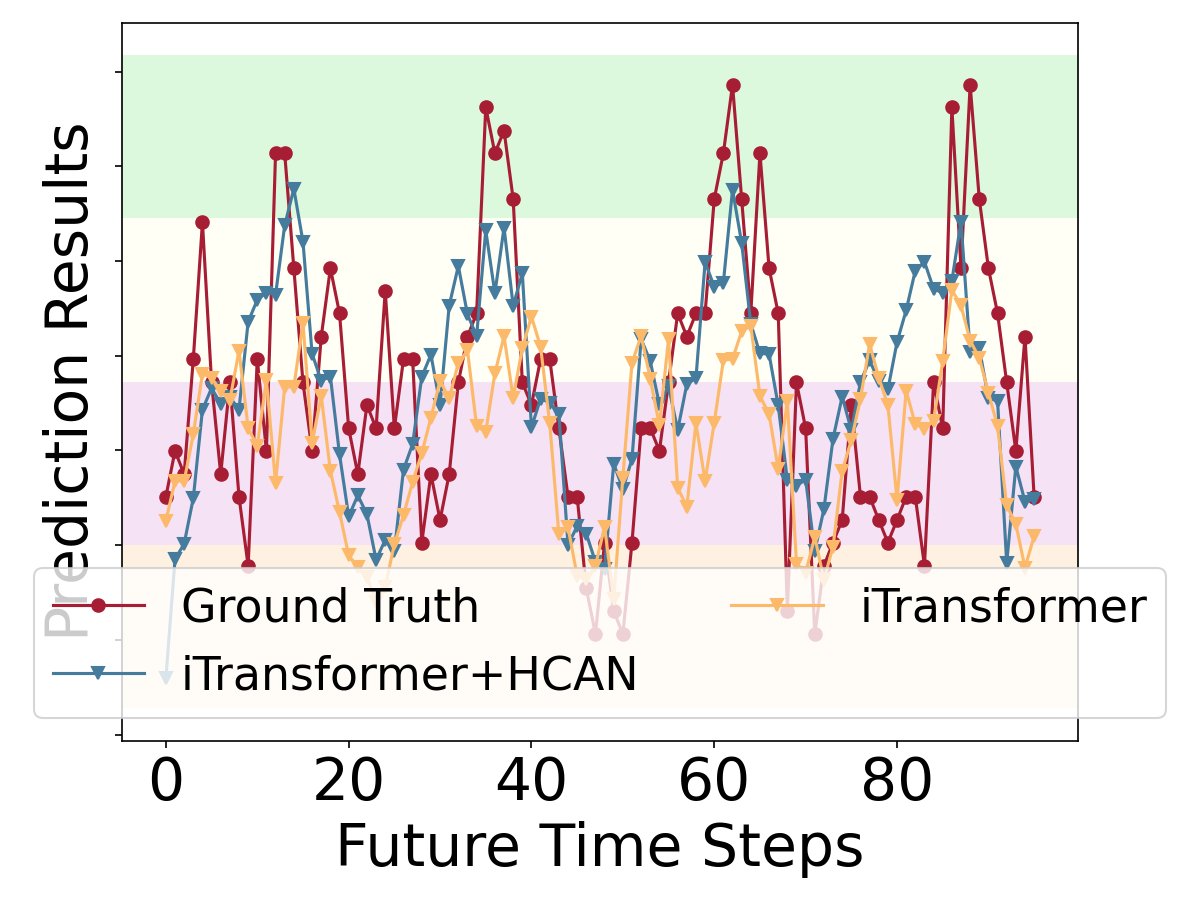}
		\caption{iTransformer+HCAN}
		\label{fig:Multivariate-iTransformer-96}
	\end{subfigure}
	\caption{The multivariate prediction results (Horizon = 96) of (a) Informer vs. Informer+HCAN, (b) Autoformer vs. Autoformer+HCAN, (c) iTransformer vs. iTransformer+HCAN, on randomly-selected sequences from the ETTh1 dataset.}
	\label{fig:Multivariate 96 results}
\end{figure*}

\begin{figure*}[!t]
\centering     
    \begin{subfigure}[t]{0.20\textwidth} 
	\includegraphics[width=\textwidth]{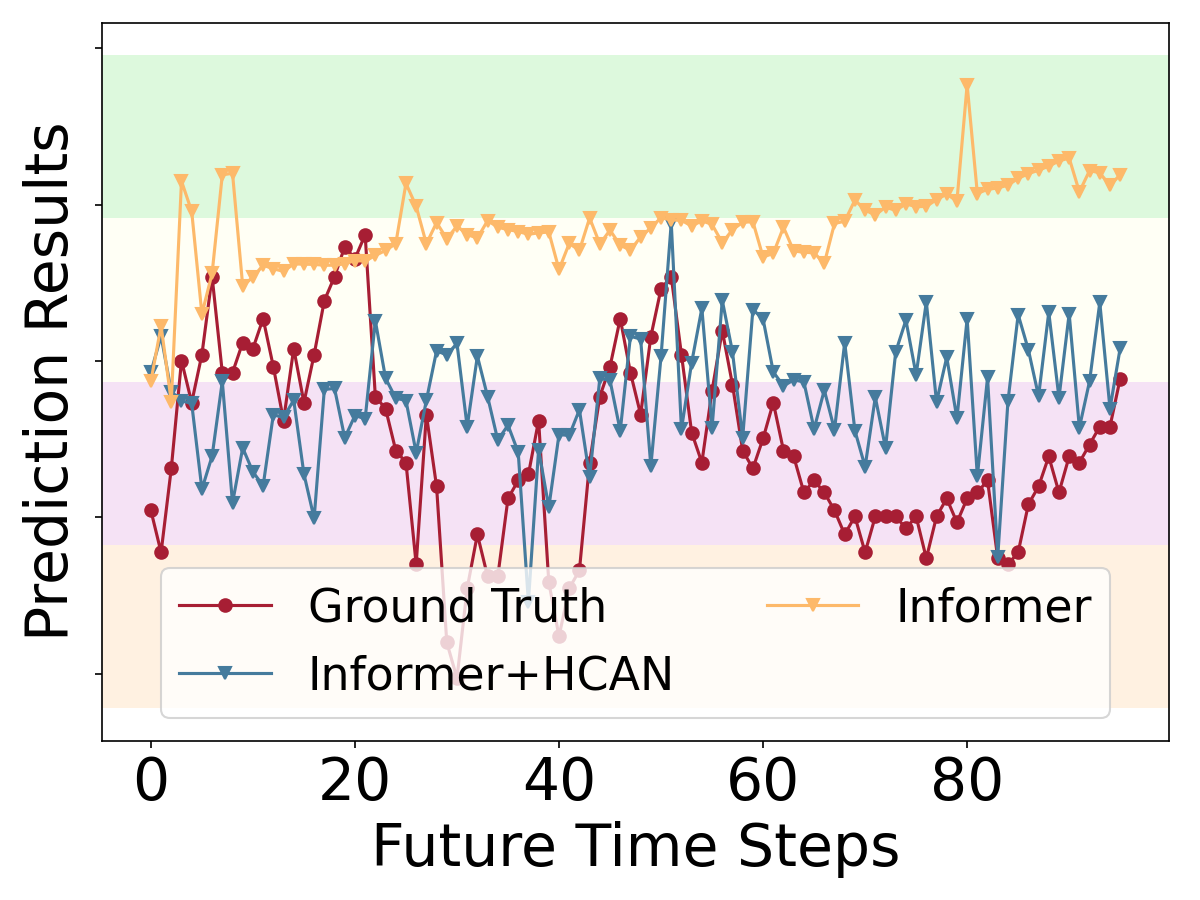}
		\caption{Informer+HCAN.}
		\label{fig:Univariate-a}
    \end{subfigure}
    \begin{subfigure}[t]{0.20\textwidth}
	\includegraphics[width=\textwidth]{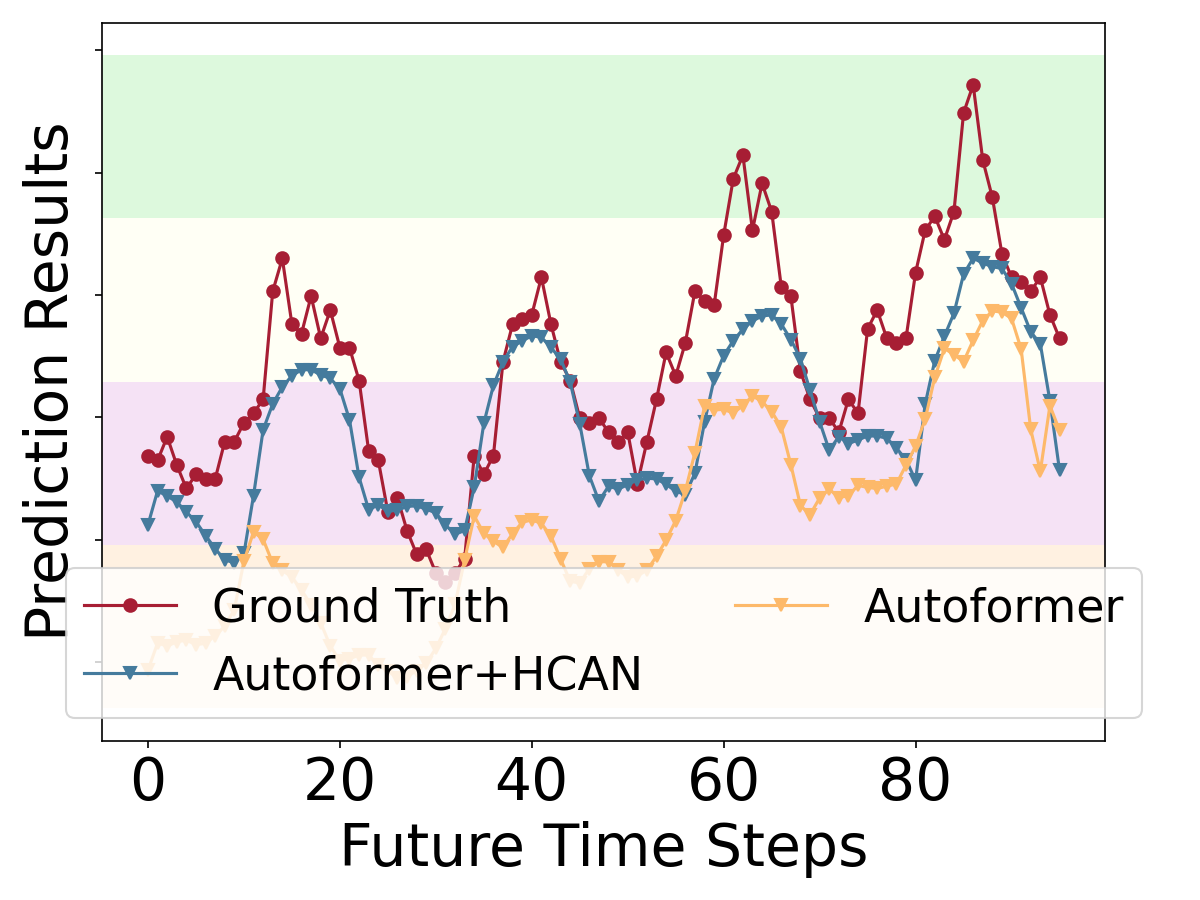}
		\caption{Autoformer+HCAN.}
		\label{fig:Univariate-b}
    \end{subfigure}
    \begin{subfigure}[t]{0.20\textwidth}
	\includegraphics[width=\textwidth]{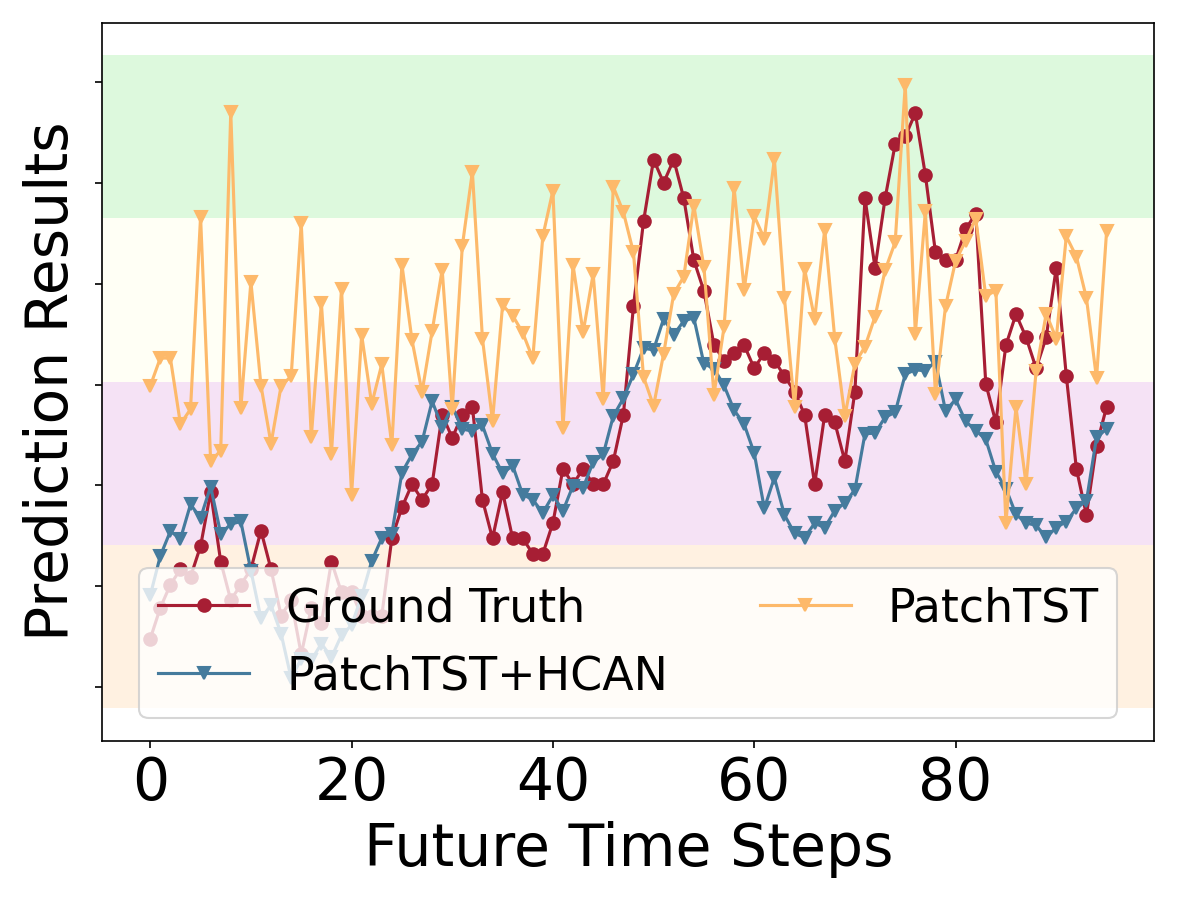}
		\caption{PatchTST+HCAN.}
		\label{fig:Univariate-c}
    \end{subfigure}
    \begin{subfigure}[t]{0.20\textwidth}
        \centering
        \includegraphics[width=\textwidth]{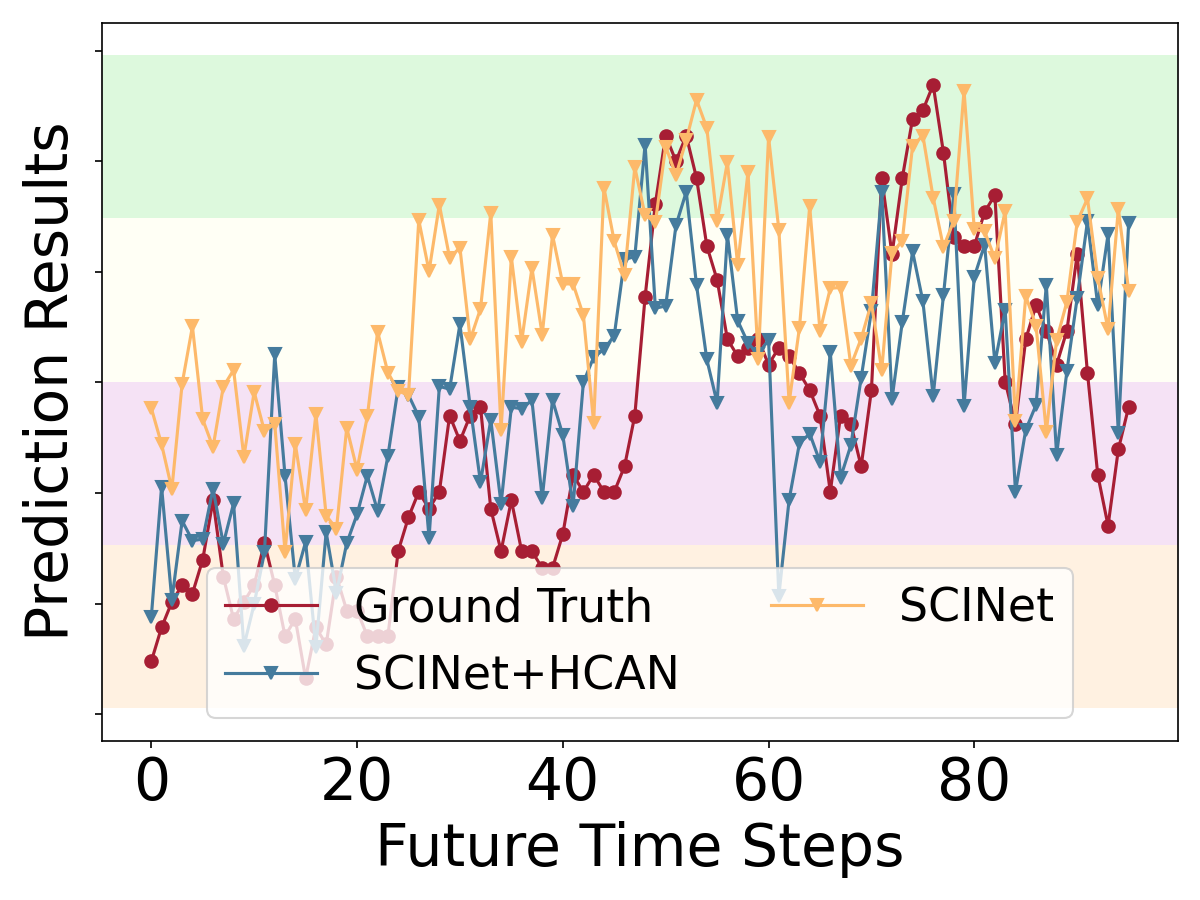}
        \caption{SCINet+HCAN.}
        \label{fig:Univariate-d}
    \end{subfigure}
    \begin{subfigure}[t]{0.20\textwidth}
        \centering
        \includegraphics[width=\textwidth]{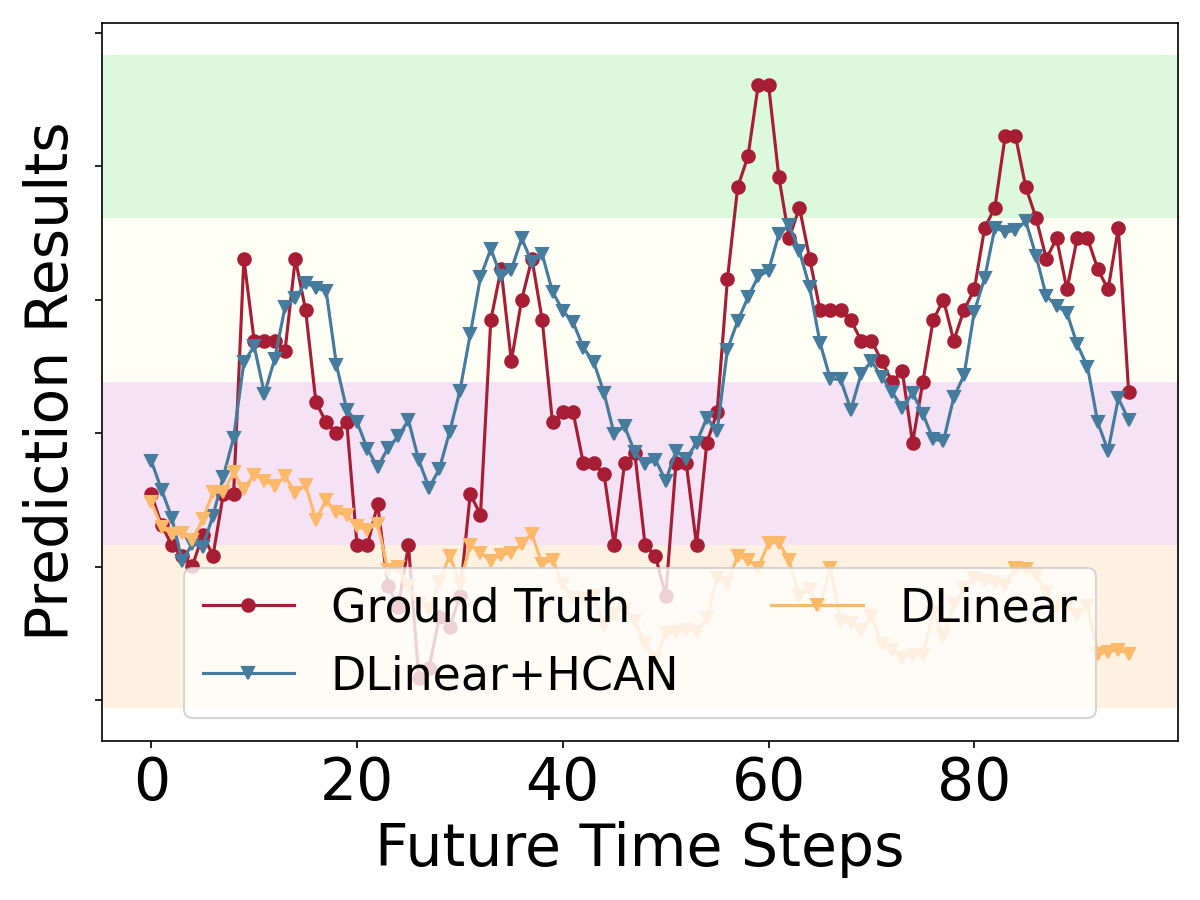}
        \caption{DLinear+HCAN.}
        \label{fig:Univariate-e}
    \end{subfigure}
    \begin{subfigure}[t]{0.20\textwidth}
        \centering
        \includegraphics[width=\textwidth]{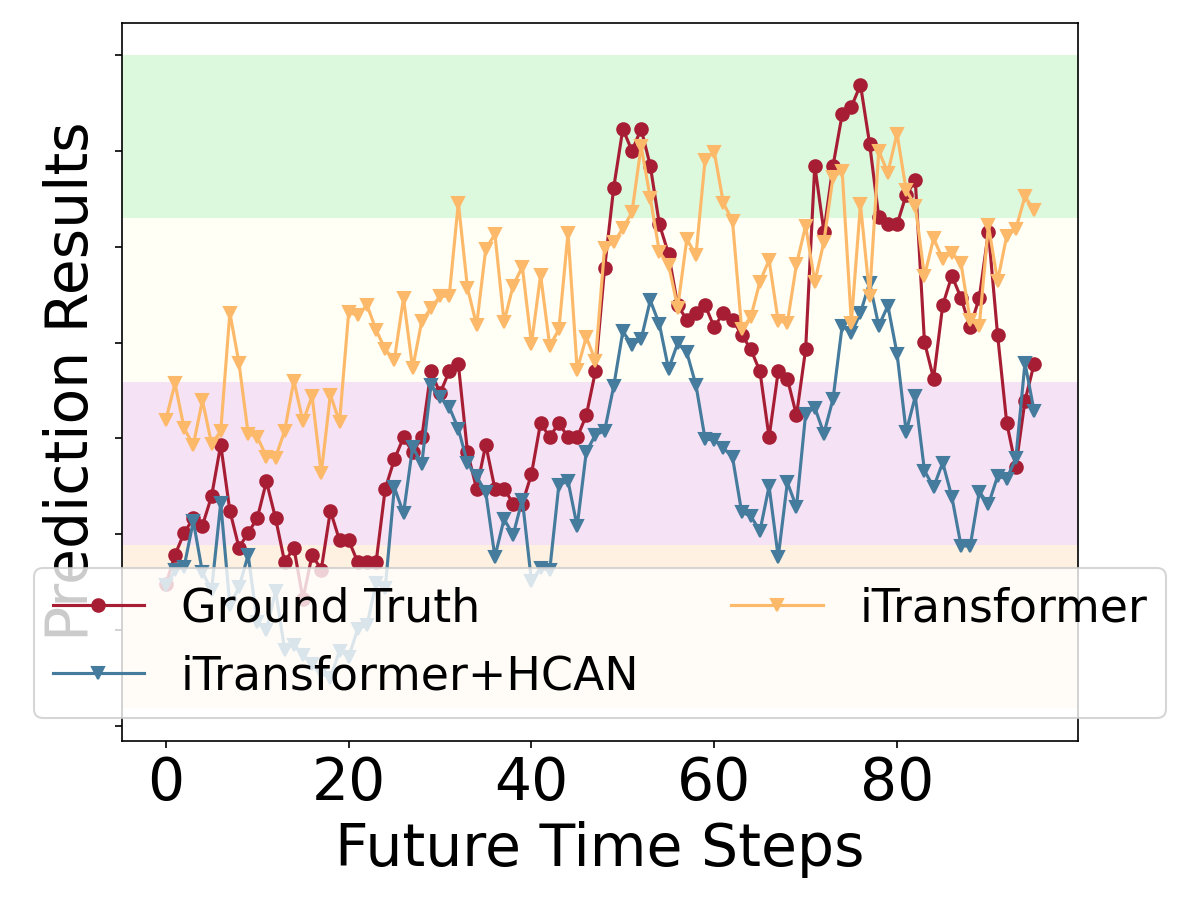}
        \caption{iTransformer+HCAN.}
        \label{fig:Univariate-f}
    \end{subfigure}
    \begin{subfigure}[t]{0.20\textwidth}
        \centering
        \includegraphics[width=\textwidth]{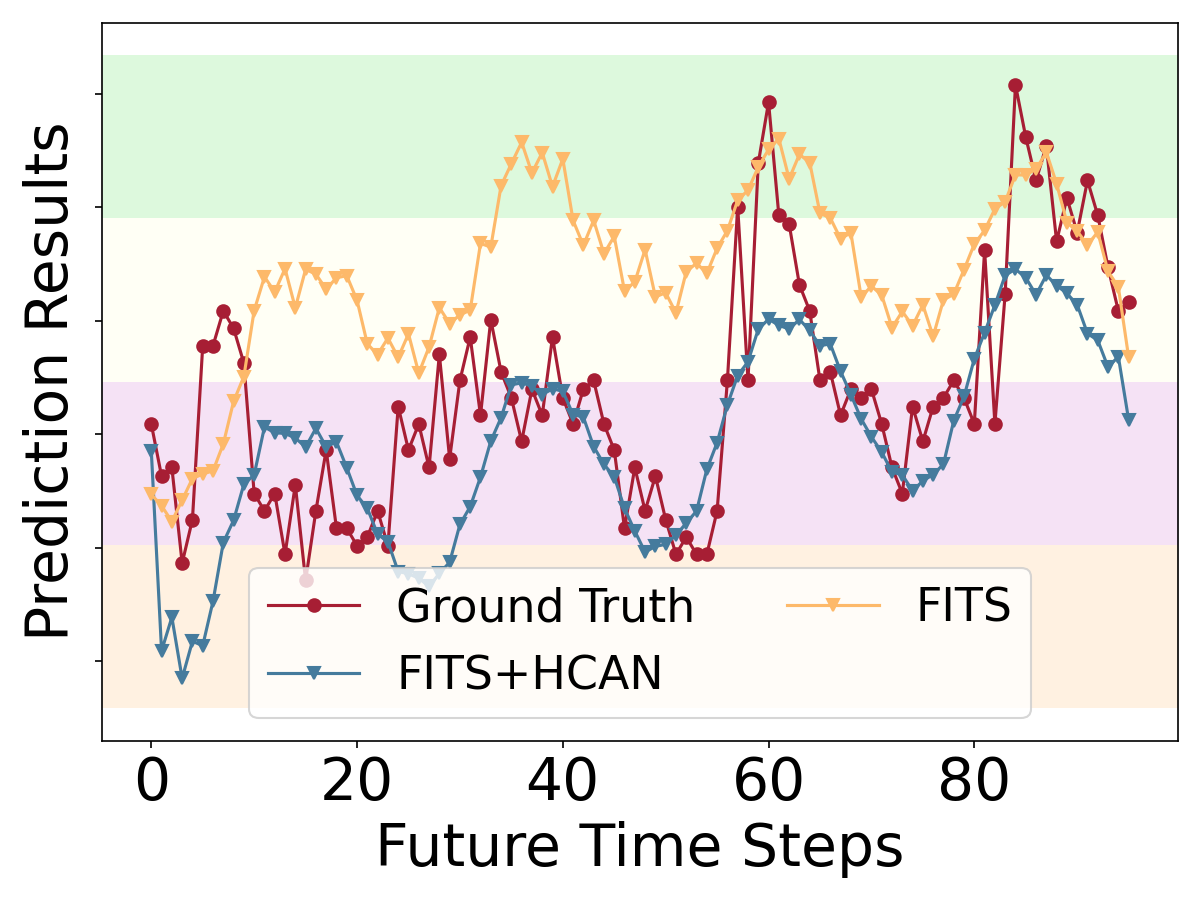}
        \caption{FITS+HCAN.}
        \label{fig:Univariate-g}
    \end{subfigure}
	\caption{The univariate prediction results (Horizon = 96) of (a) Informer vs. Informer+HCAN, (b) Autoformer vs. Autoformer+HCAN, (c) PatchTST vs. PatchTST+HCAN, (d) SCINet vs. SCINet+HCAN, (e) DLinear vs. DLinear+HCAN, (f) iTransformer vs. iTransformer+HCAN, (g) FITS vs. FITS+HCAN, on randomly-selected sequences from the ETTh1 dataset.}
	\label{fig:Univariate results}
\end{figure*}

\section{Visualizations of Main Results}
\textbf{Multivariate Forecasting Showcases and Boundary Effects.}
To evaluate the prediction of different models, Figures~\ref{fig:Multivariate 96 results} shows the comparison on Informer and Autoformer backbones on ETTh1 dataset. Similar to Figures~\ref{fig:fitting results}, the backbones tend to produce over-smooth predictions, and our HCAN gives realistic performance especially for values at the class boundaries. This is attributed to the high entropy features as well as HCL mitigating the boundary effects.

\textbf{Univariate Forecasting Showcases and Boundary Effects.}
As shown in the Figure \ref{fig:Univariate results}, adding HCAN to the baseline models gives more accurate predictions. Compared with the benchmark model, HCAN can precisely capture the periods of the future horizon by introducing hierarchical classification. In addition, our prediction series is closer to the ground truth, which can be attributed to the introduction of hierarchical attention mechanisms that enrich feature representations, along with the HCL alleviating boundary effects.

\begin{table}[!t]
	\centering
	\scalebox{0.70}{
        \begin{tabular}{ccc}
        \toprule
        \textbf{Model}   & \textbf{Number of Parameters (MB)}$\downarrow$    & \textbf{Inference Runtime (s)}$\downarrow$  \\ \midrule
        Informer                       & 43.227           & 0.0935        \\
        \rowcolor[HTML]{C0C0C0}  +HCAN & 46.273           & 0.1103        \\ \midrule
        Autoformer                     & 40.211           & 0.0474        \\
        \rowcolor[HTML]{C0C0C0} +HCAN  & 43.257           & 0.0373        \\ \midrule
        PatchTST                       & 0.134            & 0.0101        \\
        \rowcolor[HTML]{C0C0C0}  +HCAN & 3.180            & 0.0176        \\ \midrule
        SCINet                         & 0.092            & 0.1560         \\
        \rowcolor[HTML]{C0C0C0} +HCAN  & 3.138            & 0.1604        \\ \midrule
        Dlinear                        & 0.071            & 0.0006        \\
        \rowcolor[HTML]{C0C0C0} +HCAN  & 3.117            & 0.0015        \\ \midrule        
        iTransformer                   & 3.210            & 0.0026        \\
        \rowcolor[HTML]{C0C0C0} +HCAN  & 15.466           & 0.0030        \\ \midrule        
        FITS                           & 0.013            & 0.0012        \\
        \rowcolor[HTML]{C0C0C0} +HCAN  & 3.059            & 0.0027        \\ \bottomrule   
        \end{tabular}}
	\caption{Comparison of computation complexity and inference runtime.}
	\label{tab:GPU memory usage, inference time}
\end{table}

\section{Complexity and Runtime Analysis}
We compare the computational complexity and runtimes of the baseline methods with and without including our HCAN. The details are outlined in Table~\ref{tab:GPU memory usage, inference time}. Given the multi-module architecture of HCAN, it naturally exhibits a modest increase in the number of parameters and inference runtime for each model. This can be viewed as a necessary trade-off for the potential gains in forecasting accuracy and complexity handling that HCAN provides. Specifically, models integrated with HCAN, such as the Informer and Autoformer, show only slight increases in parameter size and runtime, which are offset by the enhanced capability to manage high-entropy feature representations in time series data, potentially leading to more robust and precise predictions. Additionally, the increase in inference time remains minimal, suggesting that the enhanced functionality of HCAN can be utilized with a reasonable impact on performance efficiency.

\section{Algorithm}
We provide HCAN pseudo-code based on Informer in Algorithms~\ref{alg: HCAN-Informer}. 

\begin{algorithm}
	\renewcommand{\algorithmicrequire}{\textbf{Input:}}
	\renewcommand{\algorithmicensure}{\textbf{Output:}}
	\caption{Overall Hierarchical Classification Auxiliary Network (HCAN) Procedure}
	\label{alg: HCAN-Informer}
	\begin{algorithmic}[1]
		\REQUIRE Input past time series $X$; Input Length $L$; Predict Length $T$; Data dimension $D$; Hidden dimension $M$. Technically, we set $M=512$.
		\ENSURE $\hat{Y}$, $\Delta \hat{y}_f$, $\Delta \hat{y}_c$, $e_f$, $e_c$
		\STATE F = Backbone (X) 
		\STATE $\psi$ = Linear (F) 
		\STATE $\theta$ = Linear (F) 
		\STATE $\eta$ = Linear (F) 
		\STATE $\Delta \hat{y}_f$ = Linear ($\psi$) \hfill \(\triangleright\) This is $UAC_{fine}$
		\STATE $e_f$ = Linear ($\psi$) \hfill \(\triangleright\) This is $UAC_{fine}$
		\STATE $\Delta \hat{y}_c$ = Linear ($\theta$) \hfill \(\triangleright\) This is $UAC_{coarse}$
		\STATE $e_c$ = Linear ($\theta$) \hfill \(\triangleright\) This is $UAC_{coarse}$
		\STATE $A = softmax (\psi \otimes \theta)$\hfill \(\triangleright\) This is HAA
		\STATE $\hat{Y} = Linear (Linear (A \otimes \eta) \oplus F)$\hfill \(\triangleright\) This is HAA
		\STATE \textbf{return} $\hat{Y}$, $\Delta \hat{y}_f$, $\Delta \hat{y}_c$, $e_f$, $e_c$
	\end{algorithmic}  
\end{algorithm}

\section{Broader Impact}
\textbf{Real-world applications.} HCAN addresses the crucial challenge of time series forecasting, which is a valuable and urgent demand in extensive applications. Our method achieves consistent state-of-the-art performance in six real-world applications: electricity, weather, exchange rate, illness, traffic, and space weather. Researchers in these fields stand to benefit significantly from the enhanced forecasting capabilities of HCAN. We believe that improved time series forecasting holds the potential to empower decision-making and proactively manage risks in a wide array of societal domains.

\textbf{Academic research.} HCAN draws inspiration from classical time series analysis and stochastic process theory, contributing to the field by introducing a novel framework with the assistance of hierarchical classification. The innovative HCAN architecture and associated methodologies present valuable additions to the repertoire of time series forecasting models.

\textbf{Model Robustness.} Extensive experimentation with HCAN reveals robust performance without exceptional failure cases. Notably, HCAN exhibits impressive results and sustained robustness in datasets lacking obvious periodicity, such as the Exchange dataset. The hierarchical classification structure of HCAN divides the time series into intervals for further prediction, alleviating the prediction difficulty. However, it's essential to note that, like any model, HCAN may face challenges when dealing with random or poorly temporally coherent data, where predictability is inherently limited. Understanding these nuances is crucial for appropriately applying and interpreting HCAN's outcomes.

Our work only focuses on the scientific problem, so there is no potential ethical risk.

\end{document}